\documentclass{ieeeaccess}

\usepackage{cite}
\usepackage{amsmath,amssymb,amsfonts}
\usepackage{algorithm}
\usepackage{algorithmic}
\usepackage{graphicx}
\usepackage{textcomp}

\def\BibTeX{{\rm B\kern-.05em{\sc i\kern-.025em b}\kern-.08em
    T\kern-.1667em\lower.7ex\hbox{E}\kern-.125emX}}
\usepackage{color,soul}

\usepackage{comment}
\usepackage{float}
\usepackage{newfloat}
\usepackage{listings}
\floatstyle{ruled}
\newfloat{listing}{tb}{lst}{}
\floatname{listing}{Listing}
\usepackage{color}
\usepackage{microtype}
\usepackage{tabularx}

\usepackage{amsmath,amssymb,amsfonts}
\usepackage{graphicx}
\usepackage{textcomp}
\usepackage{xcolor}
\usepackage[hidelinks]{hyperref}
\usepackage{booktabs}
\usepackage{caption}
\usepackage{subcaption}

\sethlcolor{white}

\begin{document}

\history{Date of publication xxxx 00, 0000, date of current version xxxx 00, 0000.}

\doi{xx.xxxx/ACCESS.xxxx.xxx}

\title{Evaluating Object (mis)Detection from a Safety and Reliability Perspective: Discussion and Measures} %

%Object (mis)Detection, Safety and Reliability in Autonomous Driving: Discussion, Model and Measures

% If the paper title is too long for the running head, you can set
% an abbreviated paper title here
%
\author{\uppercase{Andrea Ceccarelli}\authorrefmark{1}, 
\uppercase{Leonardo Montecchi\authorrefmark{2}}
}
\address[1]{University of Florence, Florence, Italy
 (e-mail: andrea.ceccarelli@unifi.it)}
\address[2]{NTNU, Trondheim, Norway, (e-mail: leonardo.montecchi@ntnu.no)}
\tfootnote{``This work was supported in part by the EC under Grant H2020-MSCA-823788-ADVANCE and by the National Recovery and Resilience Plan, Mission 4 Component 2-Investment 1.4-National Center for HPC, Big Data and Quantum Computing- funded by the European Union-NextGenerationEU-CUP B83C22002830001.''}

\markboth
{Ceccarelli \headeretal: Evaluating the Consequences of Object (mis)Detection}
{Ceccarelli \headeretal: Evaluating the Consequences of Object (mis)Detection}

\corresp{Corresponding author: Andrea Ceccarelli (e-mail: andrea.ceccarelli@unifi.it).}

%\institute{University of Florence, Florence, Italy \\ \and NTNU, Norway}
%******************
\begin{abstract}
We argue that object detectors in the safety critical domain should prioritize
detection of objects that are most likely to interfere with the actions of the
autonomous actor.
Especially, this applies to objects that can impact the actor's safety and
reliability. To quantify the impact of object (mis)detection on safety and reliability in the context of autonomous driving, we propose new object detection measures that reward the correct identification of objects that are most dangerous and most likely to affect driving decisions. To achieve this, we build an object criticality model to reward the detection of the objects based on proximity, orientation, and relative velocity with respect to the subject vehicle. Then, we apply our model on the recent autonomous driving dataset nuScenes, and we compare nine object detectors. Results show that, in several settings, object detectors that perform best according to the nuScenes ranking are not the preferable ones when the focus is shifted on safety and reliability.
%\keywords{Autonomous driving, safety, measures, object detection}
\end{abstract}


\titlepgskip=-15pt

\maketitle

\section{Introduction}
 %\noindent
\hl{The goal of object detection is to perceive and locate instances of semantic
objects of a certain class } \cite{jiao2019survey}. \hl{A multitude of
solutions have been proposed for $2$D and $3$D object detection,
based on cameras and lidars } \cite{liu2020deep}, \cite{zhao2019object}\hl{.
Object detection is fundamental in emerging safety-critical applications, and in particular it is a major pillar of autonomous driving applications } \cite{premebida2019rgb}.

\hl{To study object detection, new datasets are continuously
proposed, for example  in the autonomous driving domain KITTI } \cite{geiger2013vision}\hl{,  VOC } \cite{everingham2015pascal}\hl{, 
CityScapes } \cite{Cordts2016Cityscapes}\hl{, and more recently Waymo
 }\cite{ettinger2021large}\hl{, nuScenes } \cite{caesar2020nuscenes}\hl{, and Level5 Lyft }
\cite{houston2020one-level5Lyft}. Object detectors are
evaluated against datasets using widely acknowledged measures 
\cite{powers2020evaluation}, \cite{padilla2020survey},
and well-defined routines \cite{geiger2013vision}, \cite{everingham2015pascal}, \cite{Cordts2016Cityscapes}, \cite{caesar2020nuscenes}, \cite{houston2020one-level5Lyft}, allowing a fair comparison. 
\hl{Noteworthy, } the {Average Precision}, first presented in
\cite{salton1983introduction},
is currently deemed the most suitable \hl{measure} to
\hl{compute} and rank the performance of object detectors.

However, we argue that current measures for object detection 
do not match the demands and peculiarities of autonomous vehicles and safety-critical systems in general, i.e., systems whose failure may lead to harmful consequences  \cite{avizienis2004basic}.
Evaluations based on Average Precision typically judge how well a detector
detects objects, %, but they do not  consider the possible role of such objects
%within the specific scenario, and in particular, with respect to the driving
%tasks of the vehicle performing the detection.
%More precisely, the metrics describe the ability of an algorithm to detect
%objects, 
without discriminating based on the current position of these objects, and on their
possibility to interfere with the subject \hl{in the considered scenario}. % \hl{Contextualizing to  the autonomous domain,} this means that the relevance for the driving task is not considered. 
To clarify, let us consider the typical modular pipeline for autonomous driving
\cite{grigorescu2020survey}: the subject vehicle is sensing the
surroundings to perform object detection, and \hl{the output of} the object detection is used for \hl{trajectory} planning. Let us now consider two other vehicles in the sensed scenario,
one directed straight towards the subject vehicle, in a colliding trajectory,
and one headed away from the subject vehicle \hl{at a higher speed}. Clearly, for the safety of the
driving task, it is critical to detect the first one, while detection of the
second vehicle is not relevant at all. Unfortunately, this is not captured by
the measures currently used in object detection, which consider both
objects as equally relevant.\hl{ Very practically, in a typical autonomous driving modular pipeline, it is first essential to detect all relevant objects, then these objects can be used for, e.g., trajectory planning. We argue that it is desirable the object detector does not fail to detect objects in colliding trajectory, otherwise also the output of the trajectory planner is compromised.}

In this paper, we elaborate on how to measure the performance of object detectors in the
safety-critical domain, with specific contextualization to the domain of
autonomous driving, and we identify the need of an object criticality model and related measures.
As key requirement, the desired measures should reward the detection of those objects that may interfere with the subject vehicle, and %whose presence on the scene 
that are relevant for the safe and reliable execution of the driving task. %This depends on object distance, colliding trajectory, and time to possible collisions. 
Also, to be practically useful, the proposed measures have to be in a defined range, and be summarized by an overarching unifying measure. \hl{While autonomous driving is the most evident application domain, and it will be used as reference in the rest of this paper, our reasoning applies to any domain where reliability and safety of the object detection task are relevant for the success of the mission, for example in case of navigation and collision avoidance in drone systems } \cite{wozniak2022deep}.

%\hl{This paper proposes a measure to  rank object detectors based on their ability to detect objects in a colliding trajectory: we are not expecting that the object detector has the capability to predict the trajectory of the detected object, but, when the object is in a dangerous trajectory, it shall be detected. } 

%To this end, we build an object criticality model that performs a rating of the
%objects in the scene, based on the distance from the subject vehicle, the possible colliding trajectory, and the %expected time to collision. %The
%model modifies the usual metrics
%%(precision and recall, and consequently Average
%%Precision) 
%to include such rating, and revises them considering the concepts of
%safety and reliability.
\hl{More in detail, we propose a set of new measures, that we refer to as } object criticality model. Such an object criticality model assigns a criticality score to each object, based on ground truth and estimated object distance, colliding trajectory, and time to collisions. Such criticality scores contribute to compute measures, named reliability-weighted precision and safety-weighted recall, that weight correct object detections and misdetections based on the impact on the safety and the reliability of the driving task. Last, a summarizing measure, named Critical Average  Precision,  allows ranking detectors according to such safety- and reliability-oriented measures.

The object criticality model and the related measures are
exercised on the nuScenes dataset, with nine 3D-object detectors. %, all of them from the official ranking of
%nuScenes at the time of writing.  
%\fxerror{Mettere link?}
%\fxerror{Verificare i nomi dei detector}
%seven of them use lidars, while one
%relies on the visual camera.
% and we compute our proposed measure for all of them. 
We show that, under
numerous settings, the ranking we obtain differs from the one achieved using the nuScenes evaluation library, which relies on traditional measures. Amongst implications, this result questions the usual approach to rate and select the most suitable object detector for the autonomous driving domain.

The rest of the paper is organized as follows. Section II presents basic
notions and the related works. Section III shows the object criticality model and the measures we are
introducing. Section IV describes the experiments based on nine object detectors and the nuScenes dataset.
Section V illustrates the results, in which the object detectors are ranked
according to our and traditional measures, and differences are discussed.
Section VI  concludes the paper.

\section{Background and Related Works}
\label{sec:background}
 %\noindent
\subsection{Object detection and its evaluation}
\label{sec:metrics} 
 %\noindent
We report the minimal set of notions on object detection that we require to present the choices made in our work.

To describe the spatial location and extent of a detectable object, in this paper for simplicity we only consider bounding boxes, although alternative approaches, e.g., % \cite{russakovsky2015imagenet}, 
\cite{liu2020deep}, are applicable to our object criticality model as well.

Object detectors compute bounding boxes with an assigned confidence score. Then, a
\emph{detection threshold} is applied as a configuration parameter: all bounding boxes with a confidence score above the detection threshold are predictions.
%Especially when lidars' pointclouds are involved, such detectors are organized in a backbone that performs feature extraction from the input, an optional neck that further elaborates features, and an head that finally produces the predictions \cite{}.
% \subsection{Computing the confusion matrix}
%\lm{Non mi piace molto il titolo di questa sezione, ma al momento non mi viene
%niente di meglio. Anche perché mi fa un po' strano che non si usa mai il
%termine ``confusion matrix''.}\ac{Fondiamo con quella sotto? B e C insieme?}
%\lm{Ho provato a cambiare il titolo}
The classification of true positives (TPs), false positives (FPs), and false negatives (FNs), is based on some definition of distance between the predicted bounding boxes and the ground truth bounding boxes. In this paper, we use the distance between their center points \cite{caesar2020nuscenes}: a detected object is considered a TP if the distance between the ground truth bounding box and the detected bounding box is closer than a \emph{distance limit}.

\hl{If there is no predicted bounding box that matches this criterion, then the object is not detected and it counts as an FN. Predicted bounding boxes that are farther than the distance limit from all ground truth bounding boxes are considered FPs. True negatives (TNs) are not taken into account, because there are infinite bounding boxes that should not be detected within any given image } \cite{padilla2020survey}. 
% a 
% limit value is defined on the distance between bounding boxes e.g., the distance %between their
% center points \cite{nuScenes}. 
%\hl{We have a TP if the detected bounding box and the
% ground truth bounding box are closer than such limit. If there is no predicted bounding box that matches this %criterion, the object is not detected and it counts as a FN. Predicted bounding boxes that are distant from any ground %truth  bounding boxes are considered FPs.}
%(da nuscene: forse importante, ma non ho capito) In this paper we use the approach from \cite{caesar2020nuscenes}, \cite{roddick2018orthographic}, that define a match by thresholding the 2D center distance $d$ on the ground plane instead of intersection over union (IOU). This is done in order to decouple detection from object size and orientation but also because objects with small footprints, like pedestrians and bikes, if detected with a small translation error, give 0 IOU. This makes it hard to compare the performance of vision-only methods which tend to have large localization errors \cite{roddick2018orthographic}.
%\subsection{Conventional metrics}
%\label{sec:metrics}
% %\noindent
% 
% Depending on the target problem, a different set of metrics should be
% selected. This is well-known, and there are many possible examples;
% specifically,
% There is typically a set of metrics that, used together, are considered the
% de-facto standard for evaluating object detectors. These metrics are revised
% below \cite{powers2020evaluation} and are all derived from the TP, FP, FN
% concepts reviewed above.

While there are several measures that can evaluate the performance of object detectors, the conventional approach to the evaluation of object detectors consists of 
measures that are derived from the count of TP, FP, and FN.
These form the
basis for our object criticality model defined in Section \ref{sec:themodel}, and they are briefly reviewed here \cite{powers2020evaluation}, \cite{padilla2020survey}, \cite{campos2016evaluation}.
\emph{Precision}, $P= TP/(TP+FP)$, indicates how many of the selected items are relevant.\hl{ If some non-relevant items are selected, this reduces precision. Precision is 1 if all the detected objects exist, and 0 in the opposite case. }
\hl{Conversely, } \emph{Recall}, $R=TP/(TP+FN)$, %Recall indicates the number of correct detection out of all
%detectable objects. 
indicates how many of the existing relevant items are
selected. \hl{If a detector has recall 1, it means it detected everything without any detection miss; in the opposite case, recall is 0. An object detector with high recall but low precision outputs many predictions, but most of them are incorrect; an object detector with high precision but low recall returns very few predictions, but most of them are correct. }% This is captured very well by  \emph{Precision-Recall Curve}, which shows the tradeoff between precision and recall for different detection thresholds. 

Currently, the most frequently used summarizing measure is 
\emph{Average Precision} ($AP$) \cite{everingham2010pascal}, which summarizes
the precision-recall curve as the weighted mean of precision scores achieved at
different detection thresholds, using the increase in recall from the previous detection threshold as the weight. More precisely, %\cite{padilla2020survey},
%\cite{campos2016evaluation}, 
$AP=\sum_{n}{\left(R_n-R_{n-1}\right )P_n}$,
%\begin{equation}
%\label{eq:AP}
%AP=\sum_{n}{\left(R_n-R_{n-1}\right )P_n},
%\end{equation}
where $P_n$ and $R_n$ are the precision and recall at the \mbox{$n$-th}
detection threshold. %A pair $(R_k,P_k)$ is referred to as an \textit{operating
%point}.
In this paper, in agreement with \cite{caesar2020nuscenes}, we calculate
$AP$ only for recall and precision above or equal to
$0.1$: we remove cases in which recall or precision is less than $0.1$ in order to
minimize the impact of noise commonly seen in regions with low precision or low recall. % \cite{caesar2020nuscenes}.

\subsection{Related works on object detection in safety-critical systems}

A safety-critical (computer) system is
one whose malfunction could lead to unacceptable
consequences, like harm to users or to the environment. %
A typical example is an autonomous vehicle, whose malfunction (of whatever cause) may lead to a collision.  Instead, \emph{reliability} describes the continuity of correct service, which can be temporarily disrupted, for example, to avoid situations that are potentially dangerous \cite{avizienis2004basic}.

\hl{The inclusion of object detection tasks in safety-critical systems comes with a relevant set of renowned challenges, because of the many distinguishing aspects of the problem, its complexity, and also the variety of applications} \cite{amodei2016concrete},
\cite{varshney2017safety}, \cite{koopman2017autonomous}. Considering object detectors, some incorrect predictions may lead to catastrophic consequences, and therefore have the maximum impact on safety, while others may have an irrelevant impact. Further, some false positives may cause an unnecessary interruption of the service, and therefore they impact the reliability. However, to evaluate object detectors, measures from Section II.A are typically used, without considering the different impact of each detection mistake. This also applies to the wide domain of autonomous driving, and it becomes evident when considering the measures used in object detection challenges for autonomous driving. For example, in challenges for  KITTI \cite{geiger2013vision},  %\cite{kitti},
CityScapes \cite{Cordts2016Cityscapes}, % \cite{cityscapes},
Waymo \cite{ettinger2021large}, %\cite{WaymoOpen}, 
or nuScenes \cite{caesar2020nuscenes}, % \cite{nuScenes}, 
evaluation measures revolve around Average Precision and the concepts summarized in Section \ref{sec:metrics}.

Up to now, very few approaches have attempted to define safety or reliability measures for
object detectors; to the best of our knowledge, the \hl{few} works which targets a  goal \hl{similar to ours} are focusing on safety \hl{but leaving aside the reliability concern}, and are \cite{bansal2021risk}, \cite{topan2022interaction}, \cite{lyssenko2021evaluation},  \cite{Volk}. Noteworthy, they all appeared in very recent years, which underlines a recent understanding of the relevance of the subject, and they are all in the autonomous driving domain. The work in \cite{bansal2021risk} ranks each object in three categories (imminent collision, potential collision, no collision), based on its collision risk. Instead, in \cite{topan2022interaction} the authors define critical zones in which accurate perception is mandatory. On a similar position, the authors of \cite{lyssenko2021evaluation} argue the relevance of identifying a distance up to which all pedestrians are detected. The closest approach to our work is \cite{Volk}, where the authors combine scores measuring detection quality, collision potential, and time needed to make the detection. This allows computing a safety score of a test scenario, in $5$ classes from insufficient to excellent.

With respect to the reviewed works, the object criticality model we propose includes both safety and reliability issues of the driving task. This is important, because safety by itself (to detect everything which is potentially dangerous) can be enforced by low precision and high recall, i.e., low false negatives at the cost of many false positives. Instead, by balancing both reliability and safety issues, our object criticality model provides a standalone evaluation of object detectors.% we experimentally show that, when safety and reliability are central for the selection of the object detector, the most suitable one may not be the best performing according to traditional measures.

\hl{Other works address the problem of deep neural network uncertainty in autonomous driving, where the term uncertainty should be interpreted in the broad sense of how certain an object detector is about its predictions} \cite{feng2018towards}\hl{. In general, these works aim to improve  object detection, but they do not target the definition of specific measures. More specifically, the work in} \cite{feng2018towards} \hl{ arguments that object detectors should also
include prediction confidence, and it presents various methods to capture
uncertainties in object detection for autonomous driving. Otherwise, object
detectors can only tell the human drivers what they have seen, but not how
certain they are about it. The work in }
\cite{loquercio2020general} \hl{  includes information on
uncertainty sources (e.g., sensor noise), the work in } 
\cite{He_2019_CVPR}  \hl{ includes uncertainty when computing the bounding box regression loss, and the work in } \cite{kendall2017uncertainties} \hl{ considers both the noise inherent to the observations and the uncertainty that can be explained away given enough data. Last, despite not focusing on object detection, the work in }
\cite{gharib2021understanding} \hl{ defines safety-oriented measures by proposing that
predictions with a confidence score close to the detection threshold
should be treated differently and more suspiciously.} \hl{Finally, the work in } \cite{cheng2020safety} \hl{introduces the distinction of a critical area, which is the area nearby the vehicle where failed detection of an object may lead to immediate safety risks. The work acknowledges that the design of a driving application is focused on both i) guaranteeing safety in such critical area, and ii) guaranteeing high detection accuracy on the non-critical area (in order to have smooth driving). This observation leads the author to build different DNNs for the detection of objects in the two areas. }

\hl{Still, the above works weight all the detected objects the same, i.e., when assessing the object detector, the usual binary (yes/no) counting of
TPs, FPs, and FNs is performed. Instead, in our work we claim that i) object detectors should be evaluated depending on the ability to detect those objects that are most likely
to affect the driving task, i.e., impact on safety and reliability, and ii) this
can be realized by weighting the objects based on their criticality, and by
building specific measures that consider such weights. Also, we remark that, in our object criticality model, measurement errors and uncertainty in the detection are inherently considered, when computing the scores assigned to each object, and when the predicted values are compared to the ground truth.}

\section{Object Criticality Model}
\label{sec:themodel}
 %\noindent
Our \hl{object criticality} model is based on assigning a \emph{criticality} value to each object in the scene, and then computing object detection measures that consider this
criticality. The description of \hl{such} model is independent of the sensors used to
capture the scene (e.g., cameras or lidars) and of the type of objects.% in the scene.

\subsection{Requirements and assumptions}
 % %\noindent
The application of the object criticality model requires i) a subject vehicle (named \emph{ego}
afterwards) that captures the scene with sensors as cameras
and lidars, and ii) objects (other vehicles, pedestrians, etc.) that are within
line-of-sight to ego and that are consequently captured by the sensors. This is
the very typical situation of an autonomous vehicle that performs object detection.

We assume that the following ground truth information is available: i) %2D and
3D bounding boxes describing the size of the objects; ii) coordinates of ego and of the objects; iii) velocity of ego and of the objects. %This means that, to apply our model, we need autonomous driving datasets equipped with such information. 
The most recent
automotive datasets have very rich meta-data, typically including the
above information; for example, in Section \ref{sec:nuscenes} and in Section \ref{sec:results} we will use nuScenes
\cite{caesar2020nuscenes}, which satisfies our assumptions.  Clearly, the ground truth is required only to evaluate the object detector, and not in the case of operation in a deployed setting.

Further, we assume that the object detector produces as output:
i) the computed 3D bounding boxes, 
ii) the estimated distance of detected objects from ego, and
% iii) an estimate of the relative angle of objects and ego, and
iii) the estimated velocity of objects.
In other words, the object detector is assumed to conflate detection, tracking, and dynamics: this is done in several 3D object detectors, which include the above estimates in their output. Noteworthy, these estimates are computed in the object detection challenges of the nuScenes community, which will be our reference for the experiments in Section \ref{sec:nuscenes} and Section \ref{sec:results}.
% we use in this paper includes object detection challenges that require the predictions of
%the above.
% Therefore, and as such there are several object
% detectors that estimate all the above (we will use nine of them in Section V).
% We will use nine of them in our experiments in
%Section \ref{sec:results}.

For simplicity \hl{of the discussion}, when computing coordinates of 
objects and their distance from ego, \hl{in this paper} we consider only the $(x,y)$ coordinates, i.e.,  we ignore the vertical dimension.
\hl{In other words, while extending the object criticality model to the z-dimension is definitely possible, only at the cost of slightly more complex geometric computations, in the following we exclude the relative altitude of the objects and the ego. From the point of view of results, this is not an issue, because the dataset we use in this paper was collected on essentially flat lands.  Also, note that ignoring possible
vertical offsets of objects may only \emph{reduce} their distance from ego, and it is therefore a worst-case approximation.}

%\subsection{Key definitions and notation}
% % %\noindent
%We call $ego$ the vehicle that mounts the sensors and collects data from the
%environment, and we call object $B$ any other object. We do not restrict the
%type of objects, for example $B$ can be  a car,  a pedestrian, a bike, etc. Note that for $ego$ we only have %ground truth values i.e., the
%object detector does not predict its own velocity or position.
%Where needed, we indicate with the subscript $gt$ 
%when we are referring to the ground truth, i.e., the exact values. We indicate with the subscript $pt$ when we are %referring to the prediction made by the object detector. 
%When needed, we indicate with the subscript $gt$ the ground truth for the object $B$ i.e., the exact value, and we indicate with the subscript $pt$ the prediction made by the object detector of $ego$ for the object $B$. % i.e., the values predicted by the object detector for a target object $B$. In general, we expect these are 3D bounding boxes. 

\subsection{Structure of the object criticality model}
 %\noindent
%To construct our model and the final metrics on which object detectors would be evaluated, we need to compute three aspects for each object $B$ in the scene: i) the current distance between $ego$ and $B$ i.e., the distance that is measured in the captured scene, ii) the shortest distance $d$ that the $ego$ and $B$ may reach, computed based on their speed and direction, and iii) the time $\Delta t$ that $ego$ (and $B$) needs to reach such closest distance.
We call $ego$ the roving vehicle that mounts the sensors and collects data from the
environment, and we call object $B$ any other object. 
\hl{There are no restrictions on the 
type of objects, for example $B$ can be a car,  a pedestrian, a bike, etc. } Note that for $ego$ we only have ground truth values, i.e., the object detector does not predict its own velocity or position.

The construction of our object criticality model is organized in $3$ steps, which are repeated for each
object $B$ within the line of sight of $ego$, and for both the
ground truth values and the predicted values of $B$.
% , i.e., the steps are repeated for all objects that are actually present on
% the scene.

% The first step is the identification of the closest distance $d$ that $ego$ and
% $B$ can reach, and the time $\Delta t$ that $ego$ and $B$ require to reach such
% distance. These values are input to the following step, together with other
% information as current position and speed of $ego$ and $B$.

The first step (Section \ref{sec:step1}) is the analysis of the collision scenario involving $B$
and $ego$. In this step, we calculate indicators that will be later used to
define the criticality of $B$. In particular, we calculate i) the initial distance $d$ between  $ego$ and $B$, ii) the closest distance $r$ that $ego$ and $B$ would reach, and iii) the time $\Delta t$ that $ego$ and $B$ require to reach such distance.  
These values are input to the following step, together with the current position and velocity of $ego$ and $B$. 
%The analysis of the collision scenario is explored in Section \ref{sec:step1}.

The second step (Section \ref{sec:criticality}) is the calculation of \emph{criticality weights} that are
assigned to each object $B$. These are $\kappa_d$, $\kappa_r$, and $\kappa_t$, and they are based, respectively, on the three values calculated in the first step. 

These weights indicate the relevance of $B$ for the driving
task: weights are higher if it is more likely that $B$ may affect the behavior of
$ego$. Such weights are used as rewards or penalties depending, respectively, on whether the object has been detected or
missed.

The third step (Section \ref{sec:step3}) exploits the assigned criticality to construct aggregate \emph{safety and
reliability measures} that allow comparing different detectors. 
%The computation of the criticality is described in Section \ref{sec:step3}.

\subsection{Analytical characterization of the collision scenario}
\label{sec:step1}
\begin{figure}[b]
\centering
\includegraphics[width=.8\linewidth,trim={.5cm 2.3cm 1.2cm 0},clip]{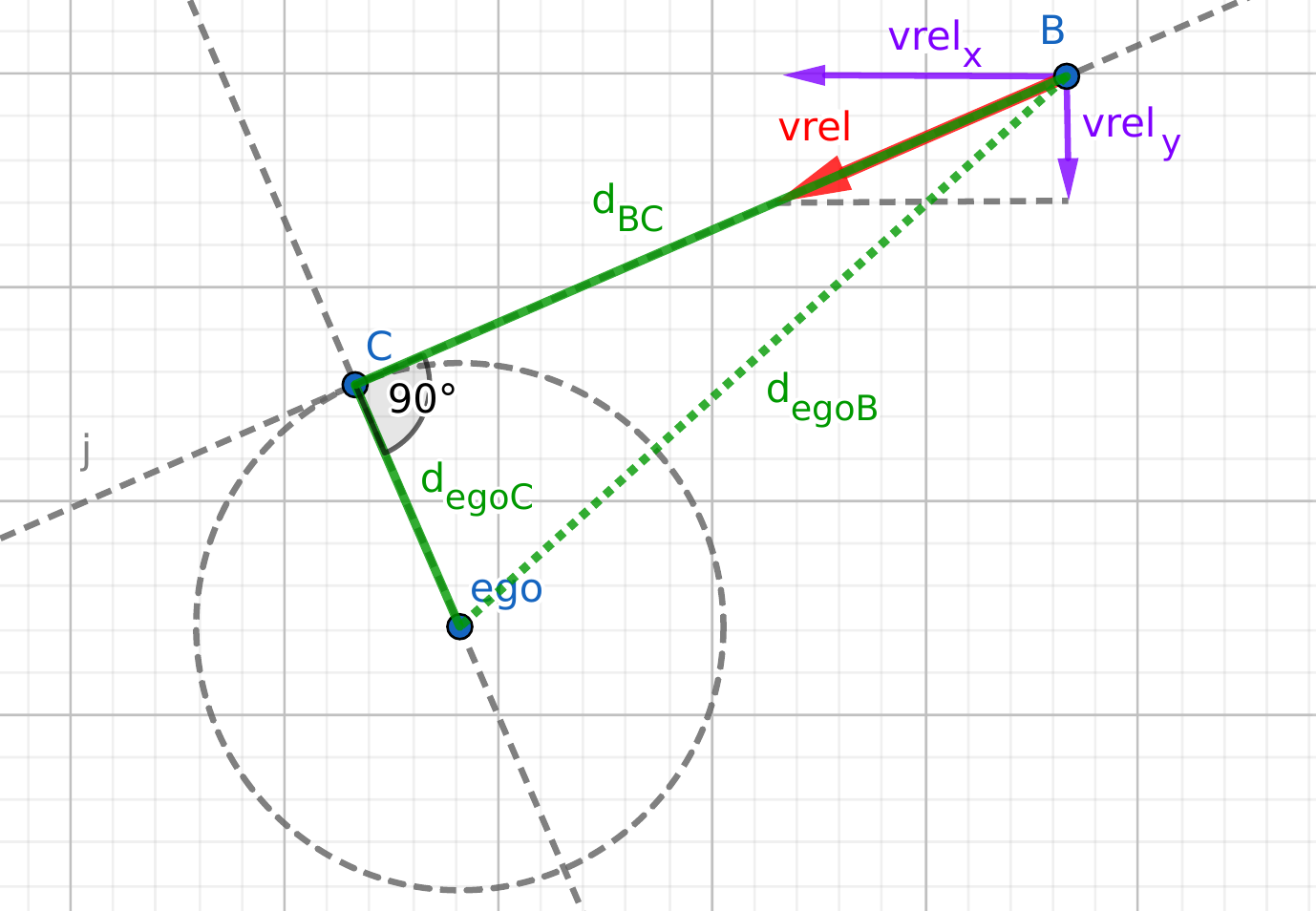}
\caption{Geometrical representation of the main elements of our object criticality model.}
\label{fig:schema}
\end{figure}

 %\noindent
We refer to Figure \ref{fig:schema} for a visual representation of the collision scenario analyzed in this section. 

We define $ego=({ego}_x,{ego}_y)$ the position of $ego$, and $B=(B_x,B_y)$ the position of the object $B$ in the captured scene. Further, we define, in vector form, $v_{ego}=(v_{ego_x},v_{ego_y})$ the velocity of $ego$, and $v_{B}=(v_{B_x},v_{B_y})$ the velocity of $B$. We compute the relative velocity of $B$ with respect to $ego$, as 
$v_{rel}=(v_{rel_x},v_{rel_y})=(v_{B_x}-v_{ego_x},v_{B_y}-v_{ego_y})$, that is, the
vectorial difference of the velocity of $ego$ and the velocity of $B$. This allows 
simplifying the subsequent calculations: we can consider $ego$ as
stationary, while $B$ is moving with the velocity resulting from the difference between the two
velocity vectors $v_{ego}$ and $v_{B}$.

Then, we identify the shortest distance from $ego$ at which object $B$ will
pass if both continue moving with the same velocity.
This is the distance between $ego$ and point $C=(C_x,C_y)$, with $C$ being
the point closest to ego on the trajectory of $B$.
% , where the trajectory is defined based on the speed vector $v_{B-ego}$. 
% That is, assuming
% that $B$ moves in a straight line, it is the point on that line that is the
% closest to the current position of $ego$.
Point $C$ can also be thought as the tangent point between the line representing
the trajectory of $B$ and a circle centered on $ego$. 
% This position is indicated by the coordinates $(C_x, C_y)$.

Point $C=(C_x, C_y)$ can be computed as the intersection of two lines, using basic Euclidean geometry. 
The line defining the direction of the relative movement of $B$ is obtained
from the general equation of a line, i.e., $y-y_0=m(x-x_0)$.
We are looking for the line passing from point $(B_x,B_y)$ and whose angular
coefficient (i.e., orientation with respect to the $x$ axis) is given by the
ratio between the $y$ and $x$ components of the relative velocity $v_{rel}$ (refer
again to Figure \ref{fig:schema}).
%That is:
%\begin{equation}
%y-B_y=m(x-B_x), \quad \text{with} \quad m=\frac{v_{rel_y}}{v_{rel_x}}.
%\end{equation}

The shortest distance between such line $j$ and the position of ego lies on the line perpendicular to $j$ passing from $ego$. By definition, a line perpendicular to
one having coefficient $m$ has coefficient
$m^\prime=-1/m$.
%To find the line that passes from the position of $ego$ is thus $y-{ego}_y=m^\prime(x-{ego}_x)$ with
%$m^\prime=-\frac{1}{m}$, that is:
%\begin{equation}
%\label{eq:sperosiagiusta}
%y=-\frac{v_{rel_x}}{v_{rel_y}}\left(x-{ego}_x\right)+{ego}_y.
%\end{equation}
%
Point $C=\left(C_x,C_y\right)$ is then the point at which these two lines intersect, which is obtained by solving Section \ref{eq:calcoloIntersezione} below. Computations are omitted for brevity.
% We find at this point the intersection of the two lines $\left(C_x,C_y\right)$.
% Point $\left(C_x,C_y\right)$ is the point at which $B$ will be closest to $ego$
% if both continue moving with the same speed and direction. 
%We omit
%computations for brevity, but the point is obtained by solving %
%Section \ref{eq:calcoloIntersezione}:
%
\begin{align}
\label{eq:calcoloIntersezione}
\begin{split}
    %\begin{cases}
      C_y=\frac{v_{rel_y}}{v_{rel_x}}\left(C_x-B_x\right)+B_y, \;\\ 
      C_y=-\frac{v_{rel_x}}{v_{rel_y}}\left(C_x-{ego}_x\right)+{ego}_y
    %\end{cases}
\end{split}
\end{align}
Then, applying the Euclidean distance, we can easily compute: 
i) the distance $d_{egoB}$ between $ego$ and $B$,
ii) the distance $d_{egoC}$ between $ego$ and $C$, and 
iii) the distance $d_{BC}$ between $B$ and $C$.%, which can be obtained simply by the Euclidean distance.

%the basic formula for the distance between two points $\alpha$ and $\beta$:
%\begin{align}
%d_{\alpha\beta}&=\sqrt{\left({\alpha}_x-\beta_x\right)^2+\left({\alpha}_y-\beta_y\right)^2}.
%\end{align}

Assuming that both $ego$ and $B$ continue moving with the same velocity, the time
$\Delta t$ that $B$ needs to reach the collision point $C$ is then computed as
the distance divided by the scalar speed of $B$, i.e., $\Delta t=d_{BC}/|v|$,
where $|v|=\sqrt{v_x^2+v_y^2}$.
We recall that $ego$ is considered to be stationary, while $B$ moves with a relative velocity obtained as the difference of the velocities of the two objects. 

We note that including acceleration would better characterize objects' movement; however, since acceleration is quadratic with respect to space, any estimation error would be greatly amplified, introducing unnecessary noise in the final measures.

Finally, note that the object criticality model exhibits some corner cases,  for example, when $ego$ and $B$ are moving at the exact same velocity, $\Delta t$ is undefined. We treat these rare cases by skipping the object criticality model calculation
and setting the criticality values to conservative fallback values. 
%Corner cases are discussed in Section \ref{sec:criticality}.  
% either the maximum (1) or minimum (0) value.

% \lm{Altre cose? Forse vale la pena dividere in subsubsections.}

% They may however happen and must be treated
% accordingly.
% 
% \ac{TODO va gestito il caso in cui Vx=0, Vy!=0, perché il rapporto diventa infinito}
% 
% \ac{TODO va gestito quando sono in coda, uno davanti all'altro, o in parallelo ma vicini (le linee non si incontrano, Vx=0/Vy=0, insomma un po' di casini). Si gestisce qua, o nella sezione dopo?}
% 
% \lm{Secondo me va bene qui. Va detto anche che si guarda se B sta andando nella direzione del punto, perché potrebbe andare nella direzione opposta. Secondo me basta fare una lista di casi particolari e come si risolvono, forse al limite anche una tabellina.}

\subsection{Computation of criticality weights}
\label{sec:criticality}

The collision scenario above is used to assign criticality to objects. The idea is inspired by reliability analysis \cite{trivedibook}, in which
quantities like {\em reliability} (or {\em safety}) are defined in the interval $[0,1]$. However, we do not propose probabilities.

Each object $B$, either identified by the object detector or
%part of %the 
ground truth, is assigned a \emph{criticality weight} $\kappa(B)$.
This weight is obtained by combining  three criticality values
$\kappa_d(B)$, $\kappa_r(B)$, and $\kappa_t(B)$, as explained later. 
Note that for a given object $B$, its criticality $\kappa(B)$ may be different if calculated with its predicted properties (e.g., position and velocity) or the ground truth ones.
Furthermore, for some objects, we may have ground truth values only (FNs) or predicted values only (FPs).
When needed, we indicate with $\kappa'(B)$ the criticality weight computed with \emph{predicted} properties of object $B$, as opposed to $\kappa(B)$ that is calculated based on the ground truth.
%We distinguish between objects that are  present in the scene, i.e., TP and FN, and objects that are not present, i.e., FP. The distinction is needed because for the first type of objects we actually have the ground truth, while for the latter we clearly do not have the ground truth because such objects do not exist, and we use the predicted values.

% We devise different types of weights, so that we can experiment with different
% approaches to the problem.\\
% \subsubsection{Approach based on threshold DISTANCE and TIME}
% In our proposal we assign a criticality weight based on both the distance
% between $d_{ego}$ and the intersection point $(C_x, C_y)$, and time $\Delta t$
% for $B$ to reach the intersection point. In the definition of our criticality
% metrics we are inspired by reliability analysis \cite{trivedibook}, in which
% metrics like {\em reliability} and {\em unreliability} are defined in the
% interval $[0,1]$. Note, however, that our metrics are not probability values.

\subsubsection{Criticality Scores.} The \emph{Distance Criticality},
$\kappa_d(B)$, is based on the distance $d_{egoB}$ between $ego$
and the object $B$. This score does not depend on velocity, but only on the position of objects in the scene. We want the score to be maximum when the distance from $ego$ to $B$  is zero, and then decrease to zero when reaching a maximum distance $D_{max}>0$.

We compute the weight $\kappa_d(B)$ as a
second-degree equation (downward parabola) passing from points $(0,1)$ and $(D_{max},0)$.
%\begin{equation}
%\kappa_d(B) = -\frac{1}{D_{max}^2} d_{egoB}^2 + 1.
%\end{equation}
That is, the maximum value is $1.0$ when $d_{egoB}=0$ and
it decreases as $d_{egoB}$ increases, reaching $0$ when $d_{ego}=D_{max}$. The parabola shape allows the
criticality to decrease non-linearly with respect to the distance: the decrease is
slow for values close to zero (i.e., close to the vehicle), and it gets faster when
approaching $D_{max}$ (i.e., far from the vehicle).
We also need to enforce that $\kappa_d(B)$ is always in the interval $[0,1]$,
and therefore the final equation is:
\begin{equation}
\label{eq:critd}
\kappa_d(B)\!=\! \max\!\left(\!0, -\frac{1}{Z^2} x^2 \!+\! 1\!\right)\!, x\!=\!d_{egoB}, Z\!=\!D_{max}.
\end{equation}

%
%\subsubsection{Collision Distance Criticality}
%
The \emph{Collision Distance Criticality},
$\kappa_r(B)$, is based on the distance between $ego$ and the
potential collision point $C$. It is an indicator of how close to ego the object is
likely to pass.
$\kappa_r(B)$ is calculated using the same rationale of $\kappa_d(B)$ (Section \ref{eq:critd}),
with $x=d_{egoC}$ and $Z=R_{max}$, where
$R_{max}>0$ is the maximum considered collision distance, beyond which the corresponding criticality is zero.

%\subsubsection{Collision Time Criticality}

Similarly, the \emph{Collision Time Criticality}, $\kappa_t(B)$, is based on the time $\Delta t$ for $B$ to reach
the potential collision point. All the other things unchanged, this score
depends on the (relative) velocity of the object $B$ with respect to $ego$.
This score is again calculated based on Section \ref{eq:critd}, with $x=\Delta t$ and $Z=T_{max}$.

\subsubsection{Object Criticality.}

The final criticality $\kappa(B)$ is obtained by the combination of the three
 criticality scores $\kappa_d(B)$, $\kappa_r(B)$, $\kappa_t(B)$. The resulting measure is defined following four 
requirements: i) it should range in the interval $[0,1]$; ii) it should be $0$
if all values are zero; iii) it should be $1$ if at least
one of the values is $1$; and iv) it should increase if any of the
three values increases.

Inspired again by classic reliability analysis \cite{trivedibook}, our final criticality weight is then computed as:
\begin{equation}
%\begin{split}
%1 - \kappa &= (1-\kappa_d)\cdot(1-\kappa_r)\cdot(1-\kappa_t)\\
\kappa = 1-(1-\kappa_d)\cdot(1-\kappa_r)\cdot(1-\kappa_t).
%\end{split}
\end{equation}
%where $\kappa(B)$, $\kappa_d(B)$, $\kappa_r(B)$, and $\kappa_t(B)$ have been abbreviated for simplicity as $\kappa$, $\kappa_d$, $\kappa_r$, and $\kappa_t$.
The final criticality $\kappa(B)$ is therefore a measure of: 
how much the object is close, 
how much it is likely to pass close in the near future, and 
how much time is available to react.

\subsubsection{Corner Cases} \hl{The following corner cases are considered:}
\begin{itemize}
  \item \hl{When \emph{ego} and $B$ are moving at the exact same velocity (in both
  dimensions), the resulting relative velocity is zero, and $\Delta t$
  cannot be computed. We solve this case by setting $\kappa_r$ and $\kappa_t$
  to zero.}
  \item \hl{The case in which only one component of the relative velocity is zero
  \emph{does not} need to be treated differently. The resolution of
  Section }\ref{sec:step1} \hl{ yields a form in which the denominator is the sum 
  of squares of the two components of the velocity. The denominator is thus
  zero only when both components of the velocity are zero, which is already treated   in the previous case.}
  \item \hl{In the calculation of $\Delta t$ we need to verify if the object B is
  actually moving towards point $C$, and not on the
  same line but in the opposite direction. In case B is moving in the opposite
  direction, $\kappa_r$ and $\kappa_t$ are again set to zero.}
  \item \hl{In rare cases, where the collision point is particularly far away or the
  speed is particularly low, the calculation of
  $\Delta t$ may generate an overflow or a not-a-number (NaN) value: in this
  case $\kappa_t$ is set to 0.1. The rationale is to set it to a low value, but
  still greater than zero.}
  \item \hl{The dataset may contain invalid values, or the detector may not be able
  to provide estimates. In particular, when we are not able to obtain the velocity of the object, we set $\kappa_r$ and $\kappa_t$ to
  1 (their maximum value).}
\end{itemize}

\subsection{Safety- and reliability-based measures}
\label{sec:step3}
 %\noindent

We exploit  the above criticality scores to 
%to
%to propose metrics that can rank
%object detectors according to principles of safety and reliability %of the driving task. We 
remodel the traditional recall and precision measures, such that they are more oriented towards reflecting the safety and reliability offered by object detectors.
% , that is expressed by, for each detectable object:
% \begin{itemize} \item $c$: current distance from the ego vehicle. Distance is
% important for example for nearby cars.
% \item $\Delta t$: the time to the collision with the detectable object. Time
% is related to speed, so it is important to understand the dangerousness of
% having a collision.
% \item $d$ is the distance to the collision point \end{itemize}

\subsubsection{Reliability}
Reliability measures the  \emph{continuity of correct service}. 
\cite{avizienis2004basic}.
 For a reliable driving task, a good object detector
should \emph{not} predict false positives that correspond to dangerous situations, because
they could lead to an interruption of the driving task. For example, false positives may cause  an
unnecessary brake; instead, the continuity of the driving mission may require considering some risks of collision as unavoidable. This clearly conflicts with safety (which aims to minimize risks), but it is widely
accepted that safety and reliability have different goals
\cite{avizienis2004basic} and may be conflicting requirements.

For this reason, we measure the reliability of the detection task 
through a revised definition of \emph{precision}. The idea is that
false positives are penalizing the continuity of the driving process, with a
greater impact the closer they are, or are likely to be, to  ego.
We weight TPs and FPs according to the criticality $\kappa(B)$ of the associated object $B$.
% . instead of simply counting the values.
In simpler words, when a non-existing object is detected, we do not add $1$ to the count of FPs, but instead we add its criticality; the same applies to TPs.

For a correctly detected object we may use the criticality computed either using the ground
truth $(\kappa)$ or the predicted values $(\kappa')$: 
%For \emph{reliability} 
we use ground truth values at the numerator, and
predicted values at the denominator. The idea is that the detector might detect
a greater criticality (denominator) than what is actually present (numerator),
which reduces  reliability of the driving task. Also, clearly we do not have ground truth values for
FPs, because those objects do not exist.

% In this case, we use the predicted values, because we want to
% estimate the extent the object detector affects the continuity of the driving
% task. Also, clearly we do not have ground truth values for FP which actually are
% non-existent objects.

 % Consequently, in this case, we use only values predicted by the object
 % detection algorithm.

%A second formulation that we explore is when we consider the ground truth for the TP , and the predicted values for the FP at the denominator. In this case, we get a value that does not satisfy the requirement of being between $[0,1]$ of precision.

We can then define the \emph{reliability-weighted
precision}  as:
\begin{equation}
{P}_{\mathcal{R}} =\frac{\sum_{B \in TP^*} \kappa(B)}{\sum_{B\in TP^*
}\kappa'(B) +\sum_{B\in FP^*}\kappa'(B)},
\end{equation}
where $TP^*$ is the set of true positive objects, and
$FP^*$ is the set of false positive objects. 
%Note that the ${P}_{\mathcal{R}}$ remains in the interval $[0,1]$ as e for precision.
Note that the ${P}_{\mathcal{R}}$ may in principle raise above 1, in case the detected
criticality is significantly lower than the ground truth. To be consistent with
the classic definition of precision, we limit the maximum value of
${P}_{\mathcal{R}}$ to 1.

\subsubsection{Safety}

% We now analyze \emph{safety}. 
Safety is instead the \emph{absence of catastrophic consequences} \cite{avizienis2004basic}.
 To ensure safety, the object detector must detect as many as possible of the dangerous objects, even at the cost of raising some false alarms. %, and there is no
%concern something more is erroneously detected. Thus, 
A safety measure should then reflect how
much of the existing criticality has been detected by the object detector. The proposed measure is adapted from the recall, using
the ground truth values at the denominator and
the detected values at the numerator. Clearly, we do not have predicted values for
FNs, which are objects that have been missed.  Therefore, we define the \emph{safety-weighted recall} as:
\begin{equation}
{R}_{\mathcal{S}}=\frac{\sum_{B\in TP^*} \kappa'(B)}{\sum_{B \in TP^*} 
\kappa(B) +
\sum_{B \in FN^*} \kappa(B)}
\end{equation}

where $TP^*$ is the set of true positive objects and $FN^*$ is the set of false negative objects.
Also for $R_\mathcal{S}$ we limit its maximum value to 1.

\subsubsection{Critical Average Precision}
The proposed criticality values depend on three parameters, namely $D_{max}$,
$R_{max}$, and $T_{max}$. We can compute ${P}_{\mathcal{R}}$ and ${R}_{\mathcal{S}}$ for different
values of these parameters, to understand their evolution when different subsets
of objects are considered.
In analogy to the precision-recall curve (see Section \ref{sec:metrics}), this allows computing several
${P}_{\mathcal{R}}$-${R}_{\mathcal{S}}$ curves, one for each combination of values
$(D_{max},R_{max},T_{max})$; consequently, we can compute the Critical Average Precision $AP_{crit}$ from each of the ${P}_{\mathcal{R}}$-${R}_{\mathcal{S}}$ curves, 
%In the following, we call $AP_{crit}$ the $AP$ computed 
based on our definitions
of ${P}_{\mathcal{R}}$ and ${R}_{\mathcal{S}}$.
% \lm{Si può dire che queste curve possono aiutare a capire il tradeoff tra
% reliability e safety per un certo detector?} \ac{si, come evidente da sezione risultati, vedi parte "Q3" di quella sezione.}

% For both precision and recall, we also try different values of $D_{max}$ and
% $T_{max}$. This allows understanding the evolution of precision and recall when
% different subsets of objects are considered.
%  This allows computing several
% precision-recall curves, one for each pair $(D_{max},T_{max})$, and
% consequently, the AP can be computed from each precision-recall curve.

% The resulting output of this step should be a set of AP values, computed from
% the precision-recall curves of the different configurations.

Depending on the driving scenario and the intended system in which the  object
detector is deployed, different values of $D_{max}$, $R_{max}$, and $T_{max}$ may be
favored.
For example, an object detector which is very good on $P_{\mathcal{R}}$ could be
safely used on a highway under low traffic conditions; but if it is
not good on $R_{\mathcal{S}}$, it should not be used in an urban scenario, where
cars may approach from different directions at essentially any angle.
%%\lm{Forse toglierei}

%The last observation is the FN-measure, to better reason on safety/reliability tradeoff. Presenting the F2 or F3 Measure, as we weight recall much more than precision, is clearly correct.

%In the same way as metrics are derived from the precision and recall, we envision the $F_{SR1}=2\cdot\frac{P_R \cdot R_S}{P_R + R_S}$ and especially the $F_{SR\beta}=(1+\beta^2) \cdot\frac{P_R \cdot R_S}{\beta^2 \cdot P_R + R_S}$, which weights safety over reliability, in our context. 

%Further, additional metrics as the average difference between $crit_{pt}(B)$ and $crit_{pt}(B)$ should be included, to analyze the difference between predicted values and ground truth values, for example by:
%\begin{equation}
%\label{eq:weigtheddiff}
%diff=\sum_{\forall B} %crit_{pt}(B)-crit_{gt}(B)
%\end{equation}

%%Last, we remark that our  model %%should not replace other metrics, but %%should be complementary to them to %%provide a complete understanding of %%the object detection performance. %We
%underline that the approach is complementary to the usage of traditional
%formulations of precision, recall and average precision, which in fact will
%still be considered in the evaluation in Section \ref{sec:results}. 
%%Further, it
%%is easy to note that our metrics can %%be reduced to the traditional ones,
%%%%by simply setting all the weights %%($\kappa$ values) to $1$.
%%\lm{E forse pure questo}

\section{Case Study on the nuScenes Dataset}
\label{sec:nuscenes}
 %\noindent
 \subsection{Datasets and selected Object Detectors}
To exercise the object criticality model, we choose the nuScenes dataset for the following reasons: i) it is very recent and extensive, forged with the latest sensor technology; ii) very recent object detectors are available; iii) it includes all the necessary information to apply the object criticality model presented in Section \ref{sec:themodel}.

%\subsection{The nuScenes object detection task}

 %\noindent
NuScenes \cite{caesar2020nuscenes} is a recent large-scale
dataset for autonomous driving that reports scenes collected from a vehicle.
%equipped with $6$ cameras, $5$ radars and $1$ lidar, all with full $360$ degrees field of view. In addition, GPS coordinates and movement dynamics from an Inertial Measurement Unit (IMU) sensor are reported. 
The dataset comprises $1000$ scenes, each being $20$ seconds long and fully annotated with $3$D bounding boxes.
% for classes of objects.
\emph{Keyframes} %(image, lidar, radar)
are sampled every 0.5 seconds; five intermediate frames are collected between keyframes.
%Driving scenes have been collected in Boston and Singapore. 
%Driving routes capture a diverse set of locations (urban, 
%residential, nature, and industrial), times (day and night), and 
%weather conditions (sun, rain, and clouds).

%\subsection{3D object detection in nuScenes}
% %\noindent
Following common practices in datasets of this kind \cite{geiger2013vision},
\cite{ettinger2021large}, nuScenes defines an object detection task and
proposes related measures to officially rank object detectors on its website. The detection task in nuScenes consists in predicting the objects at each keyframe time $t$, using sensors data collected between $(t-0.5, t]$ seconds (five intermediate frames). 
%The scene collected at time $t$ is called \emph{keyframe}, and for the purpose of detection  five intermediate frames collected in $(t-0.5, t]$ are used. 
Detectable objects are all objects within $50$ meters from ego and  with line of sight.
For each object, ground truth $3$D bounding
boxes, attributes (e.g., orientation), and velocities are provided. 
%Detectable objects are organized in $10$ classes: barrier, traffic cone, bicycle, motorcycle, pedestrian, car, bus, construction vehicle, trailer, and truck: in this paper we consider only the car class for brevity. 
A detection is
successful if the distance between the centers of the predicted and ground-truth
bounding boxes is less than a distance limit $l$; four different values of $l$ are considered, which are $ l \in \{0.5, 1, 2, 4\}$ meters. For brevity of the discussion, the only objects we consider are cars.

%The main metrics nuScenes proposes to summarize results are 
%precision, recall,
%precision-recall curve, and the Average Precision. These metrics are
%computed for each individual class and for each distance limit $l$. %To indicate a metric
%calculated with a specific $l$, we use the notation
%AP$_l$, precision$_l$, recall$_l$. 
%NuScense includes additional
%derived metrics, for example it computes the mean AP by computing the average of the %collected APs over all
%distances $l$ and classes, but we do not consider further metrics  for
%brevity of the discussion. 
%  According to \cite{caesar2020nuscenes}, operating
% points where recall or precision is less than $10\%$ are removed in order to
% minimize the impact of noise commonly seen in low precision and recall regions.
% If no operating point in this region is achieved, the AP for that class is set
% to zero. If $10\%$ recall is not achieved for a particular class, all TP errors
% for that class are set to 1.

%\subsection{Object detection algorithms}
 %\noindent
 We select nine $3$D object detectors from the zoo of mmdetection3d
\cite{contributors2020mmdetection3d}, an open-source object detection toolbox
based on PyTorch for $3$D detection. 
%which also provides downloadable weights of the trained models. 
We present the object detectors below; each detector is matched to an acronym to easily distinguish it in the rest of the paper. %As typical in object detector  \cite{}, for each detector we distinguish the backbone (i.e., feature extractors), the neck (which further elaborates feature), and the head (which produces the predictions).
%For each detector we distinguish the backbone, the neck, 
%and the head.

\emph{FCOS} \cite{wang2021fcos3d} and its evolution \emph{PGD} \cite{wang2022probabilistic} use visual cameras only. %They are monocular $3$D object detectors adapted from the $2$D detector at \cite{tian2019fcos}. 
The backbone is a pretrained ResNet101  %\cite{he2016deep}
with deformable convolutions \cite{dai2017deformable}. The neck is the Feature Pyramid Network (FPN, \cite{lin2017feature}), which generates a pyramid of feature maps. % \cite{lin2017feature}.
The head that produces final predictions (deciding on object class, location, etc.) relies on an approach similar to RetinaNet \cite{lin2017focal}, which applies shared heads to operate detection of multiple targets. PGD head also includes a branch to improve the estimation of distance depth.

The other seven object detectors (see Table~\ref{tab:objdet}) process lidar's
pointcloud and they are based on the Pointpillars \cite{lang2019pointpillars}  network.
Pointpillars is well-known both for its speed and its accuracy. It exploits an encoder that learns features on pillars (vertical columns) of the point cloud to predict $3$D oriented bounding boxes for objects. The Pointpillars network consists of three main stages: i) a feature encoder network that converts a point cloud to a structured representation, namely a sparse pseudoimage; ii) a $2$D convolutional backbone to process the pseudo-image into high-level representation, extracting the features map upon which the rest of the network is used; and iii) a detection head that detects and regresses $3$D bounding boxes. We consider seven alternatives based on Pointpillars; essentially, they
use the pillar-based method from \cite{lang2019pointpillars}
to convert the point cloud into a sparse pseudoimage, and differentiate from \cite{lang2019pointpillars} by applying
different backbones, and optionally changing the necks and heads. % The seven object detectors are reviewed in Table~\ref{tab:objdet}.
%We present each  alternative
%together with an acronym to easily distinguish them in the rest of the %paper.

\begin{table}[h!]
\centering
\caption{The seven lidar-based object detectors in use.}
\label{tab:objdet}
\begin{tabularx}{\linewidth}{lX}
\textbf{Acronym} & \textbf{Short description} \\
\midrule
FPN       & Backbone is FPN.\\
REG400    & Backbone is the REGNETX-400MF from \cite{radosavovic2020designing}.\\
REGSEC & Similar to REG400, but includes the neck SECOND \cite{yan2018second}.\\
REG1.6    & Backbone is the REGNETX-1.6GF DNN from \cite{radosavovic2020designing}.\\
SEC       & Backbone is FPN, neck is SECOND \cite{yan2018second}.\\
SSN       & As in SEC above, but it adds the shape-aware grouping heads from \cite{zhu2020ssn}.\\
SSNREG    & As in SSN, but the backbone is REGNETX-400MF from \cite{radosavovic2020designing}.
\end{tabularx}
\end{table}

\subsection{Implementation of the object criticality model}
\hl{We execute all the object detectors } on the nuScenes validation set 
\cite{caesar2020nuscenes}, which consists of 150 frame sequences of 20
seconds each, \hl{ and achieved the exact same results of their authors reported at } \cite{contributors2020mmdetection3d}\hl{. This confirms that our setup of
mmdetection3d is correct. }
%The Average Precision for the detection of cars is reported in Table \ref{tab:apnormal}. %We can easily observe that, despite we use state of the art object
%detectors, the detection capability is still very
%far from perfect, especially when $l=0.5$. %However, it is well-known that object detection in complex scenarios is still prone to several misdetections.
%\subsection{Implementation of the criticality model in nuScenes}
 %\noindent

The implementation of our object criticality model exploits the development kit of nuScenes, which is available with open-source license.
%This library is the development kit of nuScenes, and it %includes the source code
%to evaluate object detectors. 
\hl{For example, the ranking of object detectors available at the nuScenes website } \cite{nuScenes} \hl{ is computed using the code of this library, but on a different test set, whose ground truth information is not released to the public.}
We extended the development kit, to have it compute the measures from our object criticality model alongside the usual measures of the nuScenes object detection challenge. 
\hl{ We compute and plot the analogous of the precision-recall curve, but with our criticality-oriented measures ${P}_{\mathcal{R}}$ and ${R}_{\mathcal{S}}$. The resulting library is available at } \cite{our-nuScenes}. \hl{ Its usage is straightforward: it is sufficient to have a working installation of nuScenes-dev, and replace with the files in } \cite{our-nuScenes} \hl{ the corresponding files of the nuScenes-dev installation. Then, the set of results will appear enriched with our measures. }
Therefore, any object detector whose output is compatible with nuScenes can be also evaluated using our library. The library is released open source on \cite{our-nuScenes}, including tutorials that reproduce the experiments described in this paper. We used the nuScene development kit v1.1.2, and we tested for compatibility up to 1.1.7. \hl{The release at } \cite{our-nuScenes} \hl{ includes a usage example, which allows repeating our experiments from the execution of the mmdetection3d object detectors to the computation of results.}

\section{Experiments and Results}
\label{sec:results}
 %\noindent
We execute the $9$ object detectors on the \hl{dataset} previously described. We compute $AP_{crit}$,  ${P}_{\mathcal{R}}$ and ${R}_{\mathcal{S}}$ for different values of $D_{max}$, $R_{max}$, and $T_{max}$. More specifically, we consider several configurations $(D_{max}, R_{max}, T_{max})$,
with $D_{max} \in \{5, 10, \ldots, 50\}$ meters, $R_{max} \in \{5, 10, \ldots,
50\}$ meters, and $T_{max} \in \{2, 4, \ldots 30\}$ seconds. Since distance is
measured starting from the center of ego, a distance of $5$ meters  includes only vehicles very close to ego; $50$ meters instead is the maximum distance
from ego that is considered in the nuScenes object detection challenge, where
objects farther than $50$ meters from ego are ignored. Overall, this leads to
$1500$ configurations $(D_{max}, R_{max}, T_{max})$, repeated for each object detector. %The entire computation was performed on a Dell 
%Tower $7810$, and lasted approximately one week. 

\begin{table*}[h!]
\setlength{\tabcolsep}{7pt}
\caption{$AP$ and $AP_{crit}$ of car detection, for nine object detectors and $l \in \{0.5, 1, 2, 4\}$, ordered by $AP$. $AP_{crit}$ is computed with $(D_{max}, R_{max}, T_{max})$ amongst the configurations that reported the highest differences between $AP_{crit}$ and $AP$ ranking. Ranking differences are in bold.}
\label{tab:ultratable}

\centering
\subfloat[$l=0.5,\; \; (20, 20, 8)$]{
\label{tab:apcrit05}
\begin{tabular}{cll}
\toprule
\emph{Detector} & {$AP$} & {$AP_{crit}$} \\
\midrule
FCOS                 & 0.118          & 0.1995\\
PGD                  & 0.172          & 0.2711\\
SEC                  & 0.677          & 0.7534\\
FPN                  & 0.682         & \textbf{0.7618}\\
REG400               & 0.690          & \textbf{0.7616}\\
SSN                  & 0.696          & 0.7717\\
REGSEC               & 0.713          & 0.7773\\
SSNREG               & 0.717          & \textbf{0.7903}\\
REG1.6               & 0.722          & \textbf{0.7895}\\
\bottomrule
\end{tabular}
}
\subfloat[$l=1, \; \; (20, 25, 10)$]{
\label{tab:apcrit10}
\begin{tabular}{cll}
\toprule
\emph{Detector} & {$AP$} & {$AP_{crit}$} \\
\midrule
FCOS                 & 0.372          & 0.4857\\
PGD                  & 0.450          & 0.5658\\
SEC                  & 0.796          & 0.8486\\
FPN                  & 0.814          & \textbf{0.8631}\\
SSN                  & 0.818          & \textbf{0.8628}\\  
REGSEC            & 0.825          & \textbf{0.8728}\\
REG400               & 0.827          & \textbf{0.8703}\\
SSNREG               & 0.835          & \textbf{0.8806}\\
REG1.6               & 0.837          & \textbf{0.8769}\\        \bottomrule
\end{tabular}
}
\subfloat[$l=2, \; \; (10, 35, 8)$]{
\label{tab:apcrit20}
\begin{tabular}{ccc}
\toprule
\emph{Detector} & {$AP$} & {$AP_{crit}$} \\
\midrule
FCOS & 0.655 & 0.7061\\
PGD & 0.703 & 0.7382\\
SEC & 0.833 & 0.8611 \\
FPN & 0.857 & \textbf{0.8748} \\
SSN & 0.860 & \textbf{0.8691} \\
REGSEC & 0.863 & \textbf{0.8838} \\
REG400 & 0.870 & \textbf{0.8839} \\
SSNREG & 0.872 & \textbf{0.8912} \\
REG1.6 & 0.874 & \textbf{0.8837}\\
\bottomrule
\end{tabular}}
\subfloat[$l=4, \; \; (20, 25, 4)$]{
\label{tab:apcrit40}
\begin{tabular}{ccc}
\toprule
\emph{Detector} & {$AP$} & {$AP_{crit}$} \\
\midrule
FCOS & 0.804 & 0.8695\\
PGD & 0.828 & 0.8860 \\
SEC & 0.852 & 0.9057 \\
FPN & 0.872 & \textbf{0.9202} \\
REGSEC & 0.875 & \textbf{0.9205} \\
SSN & 0.876 & \textbf{0.9200} \\
REG400 & 0.884 & \textbf{0.9267} \\
SSNREG & 0.886 & \textbf{0.9293} \\
REG1.6 & 0.889 & \textbf{0.9264}\\
\bottomrule
\end{tabular}}
\end{table*}

\subsection{$AP_{crit}$ and ranking of object detectors}
 %\noindent
%With the support of Table \ref{tab:ultratable}, we discuss on $AP_{crit}$ and related ranking of object detectors.
First, we  calculate the rankings of detectors based on $AP_{crit}$ for all the 1500
configurations $(D_{max}, R_{max}, T_{max})$. Many of them produced a
different ranking with respect to the one based on  $AP$. %(i.e., the one in Table \ref{tab:apnormal}).
% The ranking  based on $AP$  (Table \ref{tab:cleandata}) and the ranking based
% on $AP_{crit}$ differ for many  configurations $(D_{max}, R_{max}, T_{max})$. In
% Table \ref{tab:invertedranking} we report the number of configurations in which
% the ranking differs from the one calculated with plain $AP$.
For example, consider $l=0.5$ and $l=4$. When $l=0.5$, the ranking calculated with $AP_{crit}$ does not match the
$AP$ ranking for $567$ out of $1500$ configurations; for each of these $567$ configurations, the
difference with respect to the $AP$ ranking is 2 or 4 positions. The whole set of object detectors may change position
with respect to the $AP$ ranking, with the exception of the detector in
the $7^{th}$ and $8^{th}$ positions which are always PGD and FCOS, respectively. For $l=4$, the ranking changes in $1425$ out of $1500$ configurations, and all the object detectors may change position, including FCOS performing better than PGD.
In Table \ref{tab:ultratable}, we compare the $AP$ and $AP_{crit}$ ranking of the nine object detectors, for exemplary configurations $(D_{max}, R_{max}, T_{max})$. Noticeably,  the object detector with the highest $AP$, REG1.6, is outperformed by SSNREG and also others when we consider $AP_{crit}$.

%We also observe how FCOS reduces its distance to the other detectors with respect to %Table \ref{tab:apnormal}, especially with $l=2$ and $l=4$. This is a
%relevant result, because FCOS, which exploits the camera instead of the lidar,
%was considered by far less performing than the other detectors according to  
%Table \ref{tab:apnormal}, while it can be considered almost on par with the other
%detectors in this case.

\begin{figure}[bt]
\centering
\includegraphics[width=0.9\linewidth]{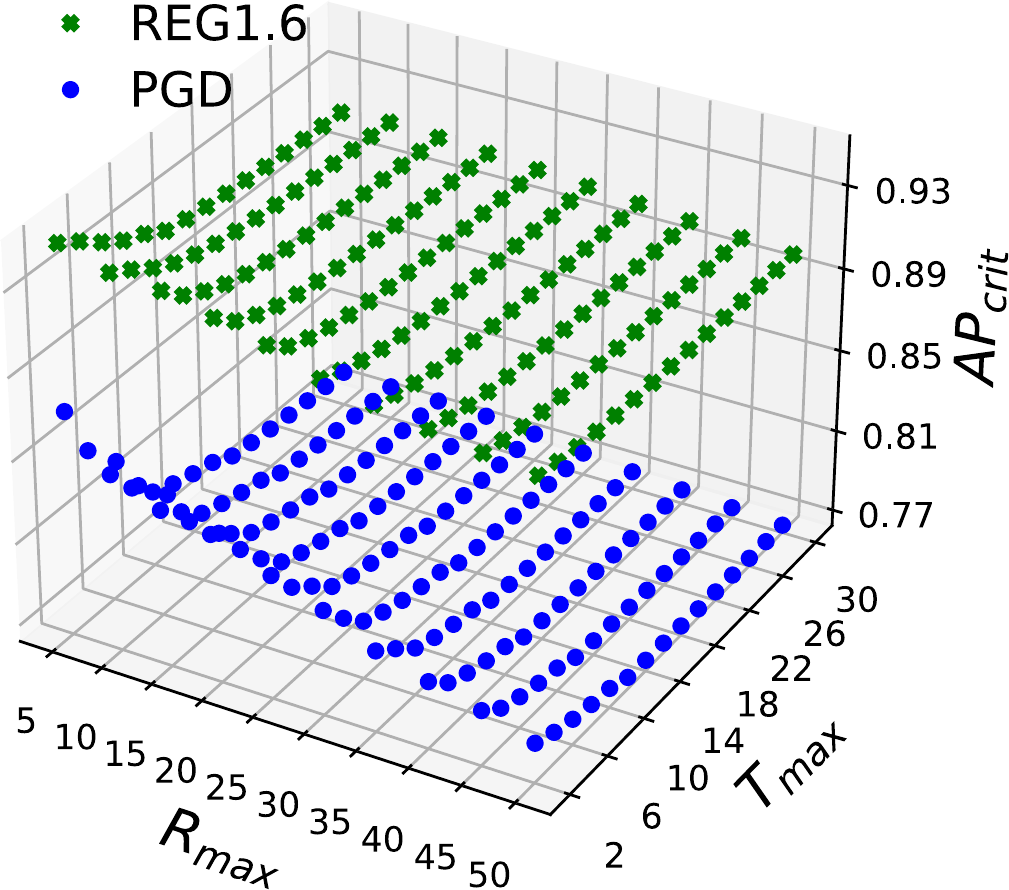}
\caption{$AP_{crit}$ measured on REG1.6 and PGD with $l=2$ and $D_{max}=25$.}
\label{fig:pianoinsalita}
\end{figure}

To explore trends of $AP_{crit}$, we select representative examples. In Figure \ref{fig:pianoinsalita} we show the  $AP_{crit}$ values of object detectors REG1.6 ($AP=0.874$) and PGD ($AP=0.703$) with $l=2$ and when $D_{max}=25$, for  different $R_{max}$, $T_{max}$. %In the lower part of the $z$ axis, 
%The corresponding $AP=0.889$ is plotted for reference; it is not affected by $D_{max}$,
%$R_{max}$ and $T_{max}$, so it is represented as a horizontal surface. 
The $AP_{crit}$ of REG1.6 and PGD is higher than the respective $AP$s under the considered
configurations. In fact, setting $D_{max}=25$ reduces the impact of objects
farther than $25$ meters, which  are
a significant contribution to misdetections.

\begin{figure}[h!]
\centering
\includegraphics[width=0.8\linewidth]{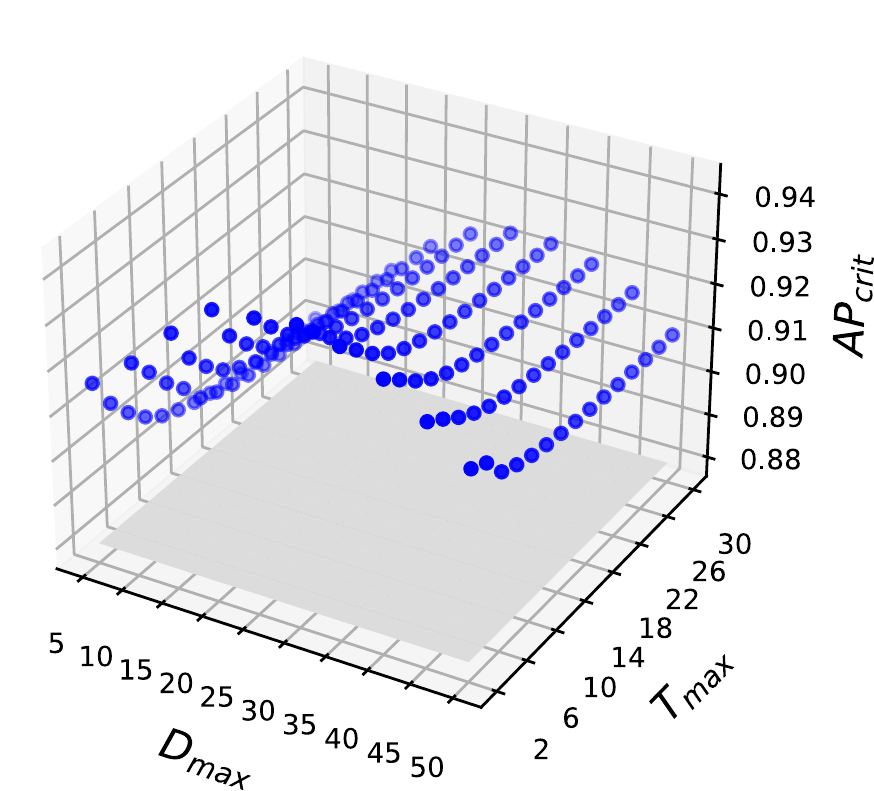}
\caption{$AP_{crit}$ measured on REG1.6 with $l=4.0$ and $R_{max}=20$, for the different $D_{max}$ and $T_{max}$.}
\label{fig:sella}
\end{figure}

\hl{Next, we pick the 
object detector REG1.6 with $l=2.0$. In Figure} \ref{fig:sella} \hl{we show the $AP_{crit}$ when $R_{max}=20$; the
figure clearly shows how the highest $AP_{crit}$ values are achieved when
$D_{max}$ is set in the range $[20, 30]$.  This is possibly due to the fact that setting $D_{max}$ very low excludes a lot of ``easy'' (i.e., close) objects from the relevant ones, thus deteriorating
$AP_{crit}$. Conversely, when $D_{max}$ becomes much greater than $R_{max}$, a
lot of distant but not relevant objects are included, which are unlikely to
reach a collision point closer than $R_{max}$. In the lower part of the $z$ axis, $AP=0.874$ is represented as a flat grey surface in the figure. Figure } \ref{fig:sella} \hl{ shows that $AP_{crit}$ is in general higher than $AP$. This is expected, because the $AP_{crit}$ gives less weight to objects that are harder to detect, e.g., those at a farther distance from ego. }%In fact: i) $D_{max}=5$ rewards only the detection of vehicles that are exceedingly close, and ignore the others, and ii) $R_{max}=50$, $T_{max}=30$ exclude the peculiarities of this model, because they are large values that introduce small or no
%$AP_{crit}$ penalizations on most of the vehicle with respect to the $AP$ case.

In general, higher values of $AP_{crit}$ are
achieved with low values of $R_{max}$ and $T_{max}$; for both REG1.6 and PGD, % (see Figure \ref{fig:pianoinsalita}), where 
the maximum $AP_{crit}$ values are obtained with  $(D_{max}, R_{max}, T_{max})=(25, 5, 2)$. Intuitively, low $R_{max}$
and $T_{max}$ reduce the number of vehicles to be considered in our analysis:
only those that are really critical for the detection are included. Analogous observations can be derived with the other configurations and object detectors.

We remark that, while studies like Figure \ref{fig:pianoinsalita} and Figure \ref{fig:sella} are effective to explain the proposed $AP_{crit}$ measure, the most suitable configuration $(D_{max}, R_{max}$, $T_{max})$ should be decided based on the requirements of the target application, and then the object detector with the highest $AP_{crit}$ for such configuration should be selected.

%The maximum $AP_{crit}$ values of  Table \ref{tab:apcrit} are obtained with the triple $(D_{max}, \\
%R_{max}, T_{max})=(25, 5, 2)$,
%with just some exceptions: the two occurrences in bold are obtained with
%$(30, 5, 2)$, and the two underlined occurrences  are
%obtained with $(20, 5, 2)$.  This does not mean that the configuration
%$(25, 5, 2)$ is the ``best'' one:
%in fact, this configuration sets strong constraints on the objects that are interesting for
%detection. We believe the entire set of configurations
%needs to be studied, and the choice of the best suited configuration should be
%based on considerations on the $AP_{crit}$ measured at $(D_{max}, R_{max},
%T_{max})$, the target system, and the expected scenario in which it will be used.
%The maximum difference between $AP$ and $AP_{crit}$ is
%$\max(|AP_{crit}-AP|)=0.289$, obtained for FCOS with $l=1$  and $(D_{max}, R_{max},
%T_{max})=(20, 5, 2)$: in this case, we have $AP_{crit} = 0.661$, while $AP=0.372$. 
%Raising $l$ corresponds to smaller differences between the
%object detectors. This corresponds to the higher differences between $AP_{crit}$
%and the corresponding $AP$ with higher $l$ values. %\lm{Non lo vedo questo aumento di
%differenza tra i detector aumentando $l$}

\subsection{Tradeoff between ${P}_{\mathcal{R}}$  and ${R}_{\mathcal{S}}$}

\begin{figure}[b!]
\centering
\includegraphics[width=0.45\textwidth]{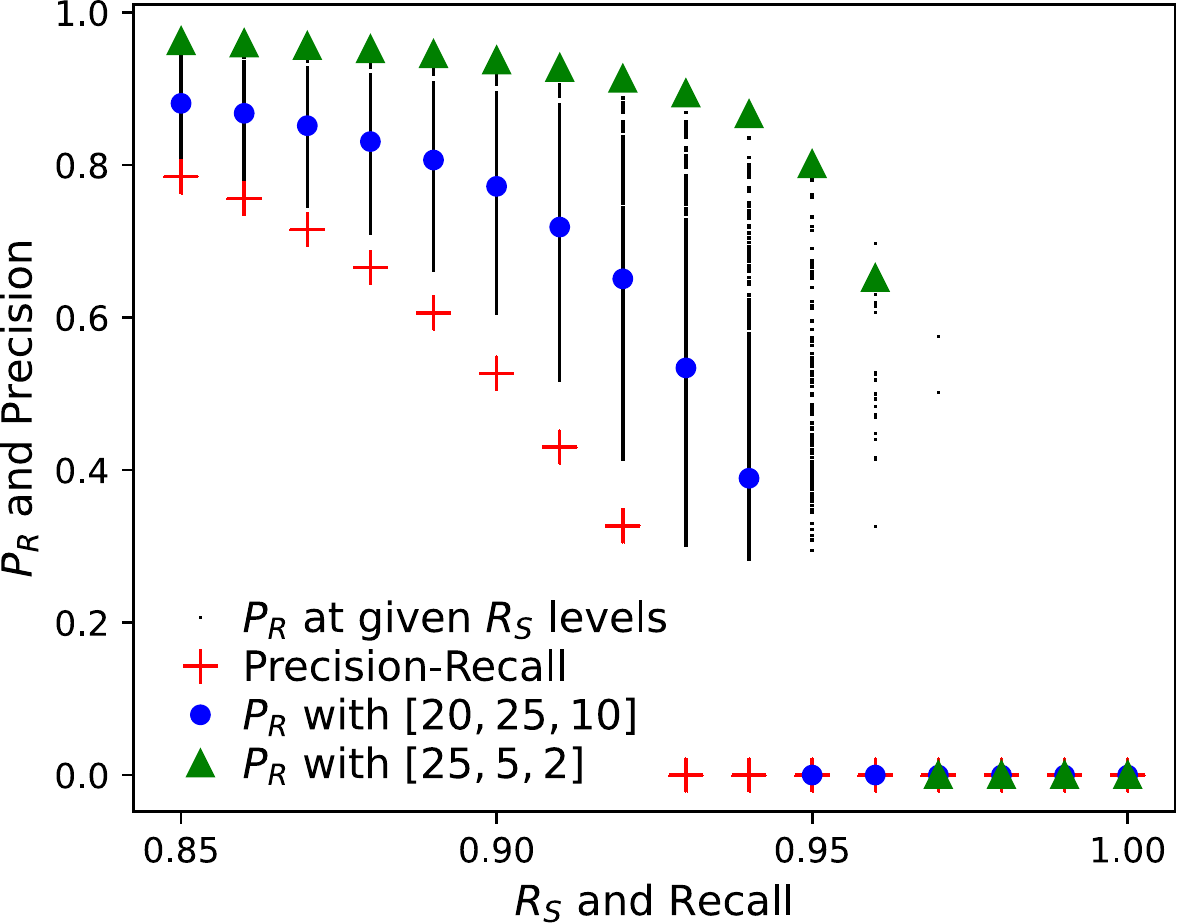}
\caption{${P}_{\mathcal{R}}$, ${R}_{\mathcal{S}}$, $P$ and $R$ for SSNREG when $R_S \geq 0.85$ and $R\geq 0.85$, with $l=1$.}
\label{fig:WTFisThis1}
\end{figure}

\begin{figure}[t!]
\centering
\includegraphics[width=0.45\textwidth]{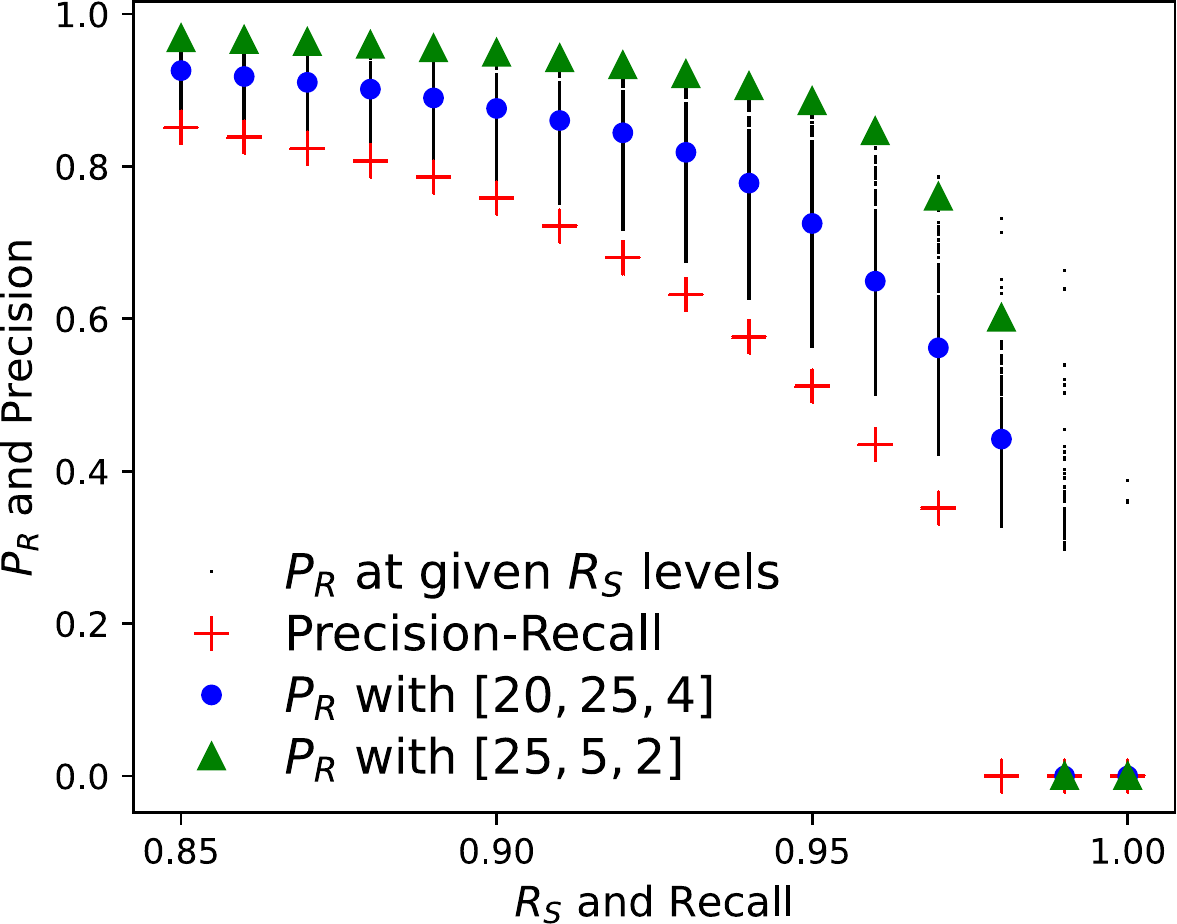}
\caption{${P}_{\mathcal{R}}$, ${R}_{\mathcal{S}}$, $P$ and $R$ for SSNREG when $R_S \geq 0.85$ and $R\geq 0.85$, with $l=4$.}
\label{fig:WTFisThis2}
\end{figure}

To discuss the relations between ${P}_{\mathcal{R}}$  and ${R}_{\mathcal{S}}$, we rely on \autoref{fig:WTFisThis1} and \autoref{fig:WTFisThis2}, where we use SSNREG with $l\in\{1, 4\}$. We compute ${P}_{\mathcal{R}}$ and $P$ for, respectively, ${R}_{\mathcal{S}}$ and $R$ at steps of $0.01$, starting from $0.85$. 
Red crosses represent precision-recall pairs $(P, R)$. Black dots represent  $(P_\mathcal{R}, R_\mathcal{S})$ pairs; these are computed for each configuration 
$(D_{max}, R_{max}, T_{max})$, thus yielding 1500 black dots for each
${R}_{\mathcal{S}}$ value.
The large blue dots are the $(P_\mathcal{R}, R_\mathcal{S})$ values achieved using 
SSNREG with the configuration from \autoref{tab:apcrit10} and \autoref{tab:apcrit40}, while the green triangle are the configuration leading to the highest $AP_{crit}$, which is $(25, 5, 2)$. % for both figures.  %The green triangles with $l \in \{2\}$ instead represents respectively the configuration  $(5, 20, 2)$ and $(5, 15, 2)$, which have the highest ${P}_{\mathcal{R}}$ when ${R}_{\mathcal{S}} \geq 0.98$. 
%We notice that higher $l$ values lead to higher values of ${P}_{\mathcal{R}}$ given a target ${R}_{\mathcal{S}}$. This is expected: with a higher $l$ value, the number of correct detections increases.

%\begin{figure}
%\centering
%\includegraphics[width=\linewidth]{img/precreccurve}
%\caption{Relations between ${P}_{\mathcal{R}}$ and ${R}_{\mathcal{S}}$ for REG1.6 with $[D_{max}, R_{max}, T_{max}]$ set to, from left to right: $[10, 10, 4]$, $[15, 15, 8]$, $[25, 25, 12]$.}
%\label{fig:WTFisThis}
%\end{figure}

%More importantly, 
We  investigate the relations between ${P}_{\mathcal{R}}$ and ${R}_{\mathcal{S}}$ for high values
of ${R}_{\mathcal{S}}$ (safety-weighted recall), which are of particular interest in the
reference domain of this work. This way we can study the ${P}_{\mathcal{R}}$ (reliability-weighted precision) that we achieve
when safety is enforced thanks to a high ${R}_{\mathcal{S}}$. 
This corresponds to answering the question
``given a safety target
on the detection, what is the possibility of driving the car
with good mission reliability, i.e., without being forced to
interrupt the driving continuously because of false positives?''. Of course, the safest condition would be $R_S=1$,
but ${P}_{\mathcal{R}}$ is typically $0$ in such cases; still, a very high ${R}_{\mathcal{S}}$ is
necessary to enforce safety of the detection. 

When the recall $R$ increases, the precision $P$ quickly drops to $0$. %, and it is always zero in the case of \autoref{fig:WTFisThis2a} where $l=0.5$. 
SSNREG can offer a high recall, i.e., a high ability to detect all the objects, only at the cost of many false
positives: this is clearly of little or no use in practice. Instead, if we
restrict the scope of the object detector thanks to our object criticality model, we
reach different conclusions. For example, consider again the case $l=1$ (\autoref{tab:apcrit10}). Even with $R_\mathcal{S} \geq 0.9$, there are some configurations in which $P_\mathcal{R}>0.8$, which is clearly a much more comforting result, showing confidence in the detection at least to some extent. 

On the other hand, the best-performing triples, represented with the green triangles in \autoref{tab:apcrit10}, may be not practical, because it is computed applying small spatial and temporal distances of the objects from ego. Summarizing, our conclusion on SSNREG can be very different from those we achieve using $P$ and $R$, when we %rate the bounding boxes to be examined (both the predicted and ground truth
%ones) to match 
apply the criteria of ${R}_{\mathcal{S}}$ and ${P}_{\mathcal{R}}$.

%%%%%%% VERIFICARE
%
%\begin{figure}
%\centering
%\includegraphics[width=\linewidth]{img/surpisingdmax}
%\caption{to be commented.}
%\label{fig:WOW}
%\end{figure}
%\subsection{Q3}
%
%\begin{figure}
%\centering
%\includegraphics[width=\linewidth]{img/8curveD25}
%\caption{to be commented.}
%\label{fig:WOW}
%\end{figure}

%\section{Limitations of our Work}
%\label{sec:limitations}
%Non so, da riempire se ci vengono idee
%\lm{Ma invece fare un ``Limitations and Future Work''? E lasciare un Conclusion
%più standard? Magari aiuta a riempire questa sezione}
%Non abbiamo limitazioni, solo lavori futuri!!!

\subsection{Explanation of distance criticality $\kappa(B)$}\hl{The objective of this analysis is to explain the inner details of the object criticality model, even if  $P_\mathcal{R}$, $R_\mathcal{S}$, and $AP_{crit}$ are sufficient to describe the performance of the object detection.  We rely on bird-views from selected frames of nuScene to  explain how our object criticality model works, in a very practical way, for the computation of $\kappa(B)$. We consider PGD and SEC object detectors, but all nine detectors lead to similar conclusions.}

\begin{figure}[h]
\centering
         \includegraphics[width=0.35\textwidth]{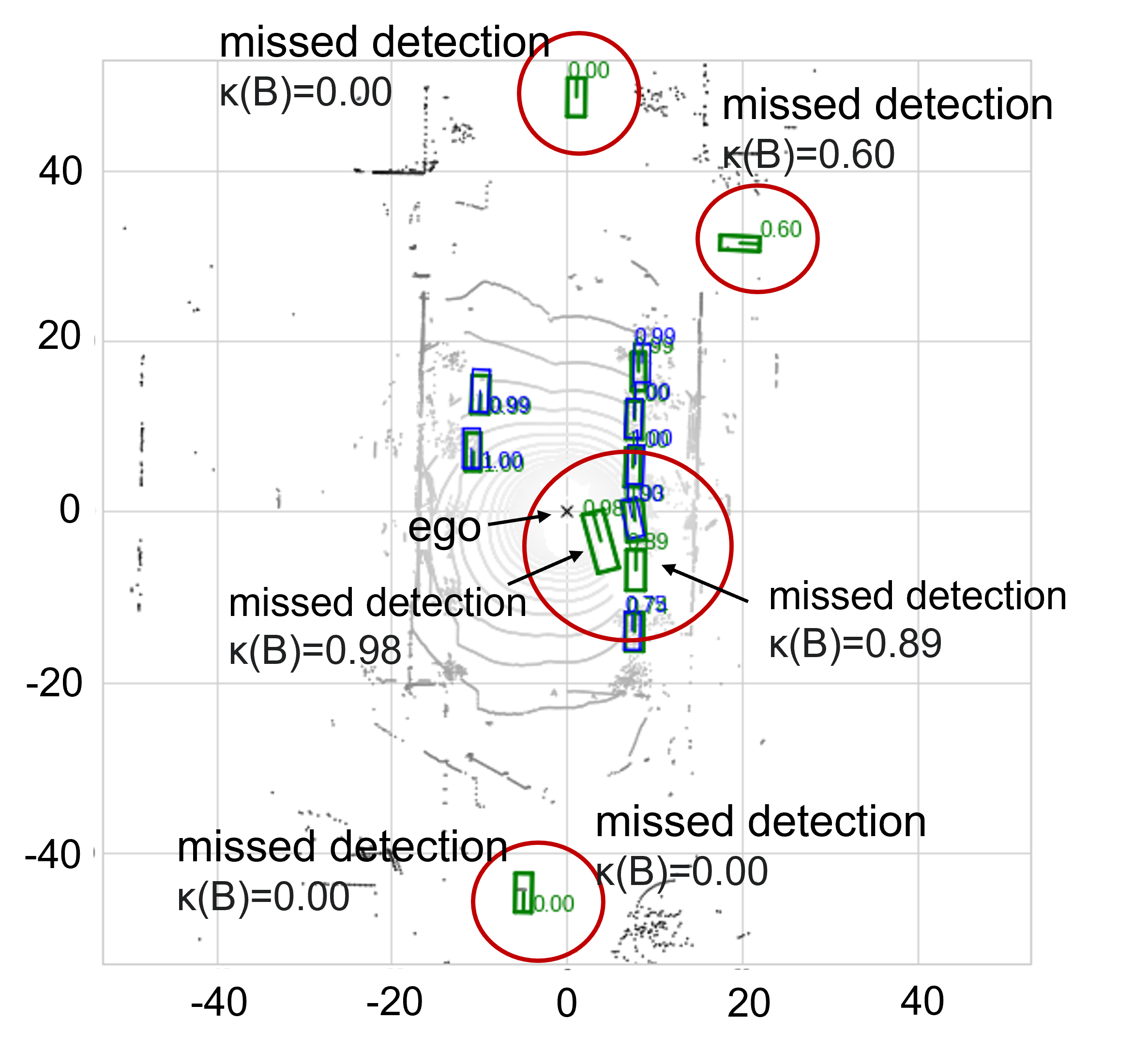}
         \caption{$\kappa(B)=0.98$ and $\kappa(B)=0.89$ for two dangerous missed detections from PGD. \textit{Best viewed in color.}}
         \label{fig:birdview1}
\end{figure}

\begin{figure}[h!]
\centering
         \includegraphics[width=0.35\textwidth]{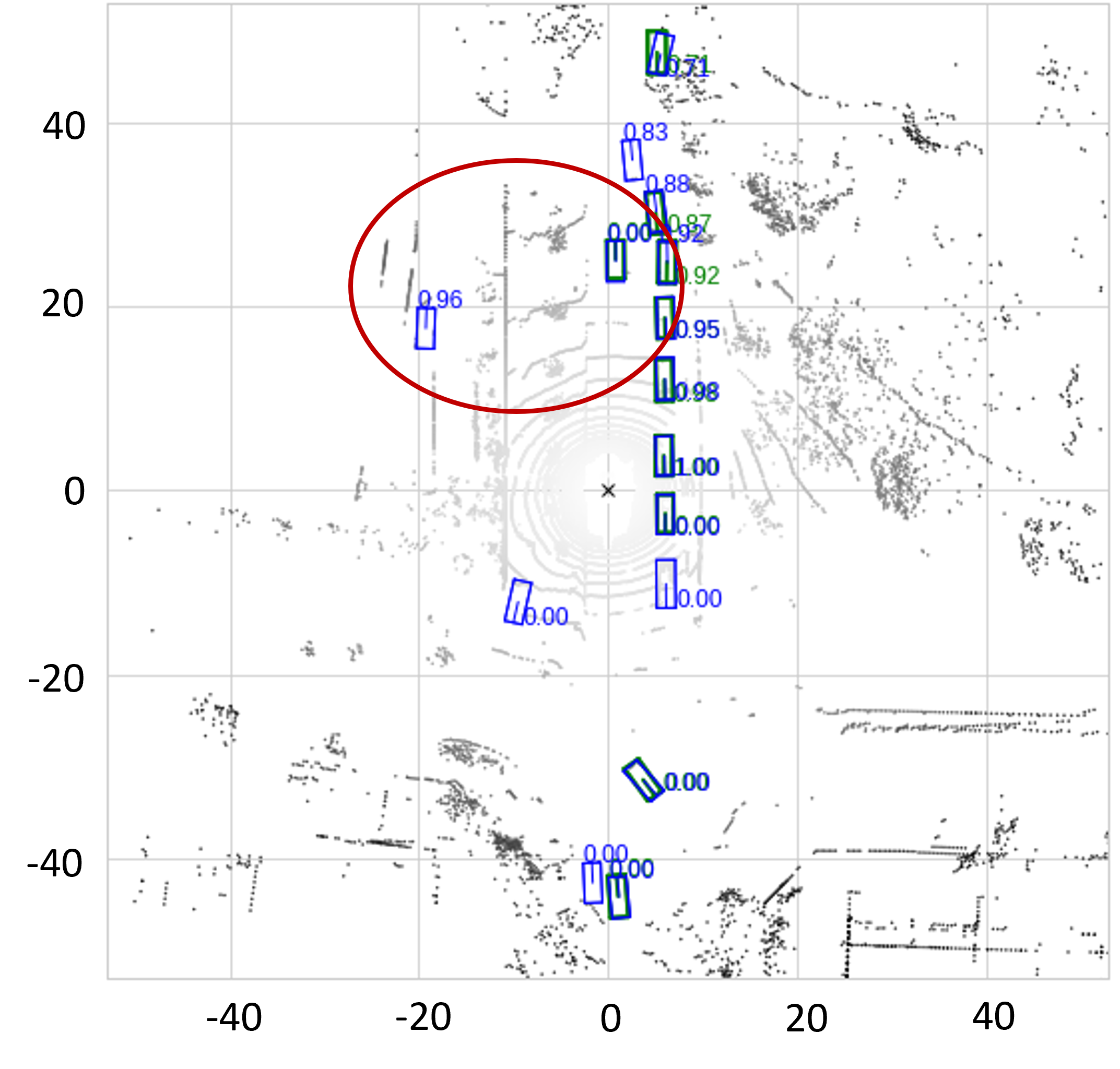}
         \caption{ $\kappa_t(B)$ computed for SEC with $D=15$, $I=20$, $T=10$. Objects which may collide with ego within $10$ seconds have assigned $\kappa_t(B) \geq 0$. \textit{Best viewed in color.}}
         \label{fig:birdview8}
\end{figure}

\hl{Figures from Figure } \ref{fig:birdview1} \hl{to Figure} \ref{fig:birdview8} \hl{are extracted relying on the nuscene-dev kit 1.1.2, properly modified to visualize values from  $\kappa(B)$, $\kappa_d(B)$, $\kappa_r(B)$, and $\kappa_t(B)$. The axes represent distances, in meters. The ego is always located in the center at the $(0, 0)$ coordinates and is oriented along the y-axis (heading towards the top). The other vehicles are represented as rectangles, and the front side is indicated by  a small segment. The ground truth (real position and orientation of cars) is in green, while the detected cars are in blue. In the ideal case of a perfect object detectors, blue and green rectangles would overlap. Both ground truths and detected vehicles have associated a value, which is either $\kappa(B)$, $\kappa_d(B)$, $\kappa_r(B)$, and $\kappa_t(B)$ depending on the figure. We add text labels and red circles to improve readability.}

\hl{We first consider PGD with $D_{max}=30$, $R_{max}=20$, $T_{max}=8.0$. Very intuitively, this setting says that it is critical to detect vehicles that are within $30$ meters, and/or that are in colliding trajectories within $20$ meters in the next $8$ seconds.}

\hl{We start from Figure } \ref{fig:birdview1}\hl{. A car is very close to ego, but it is not detected: it has been assigned $\kappa(B) = 0.98$. This car is located at the center of the diagram, and it is circled in red. Another one is very close, but in ``a less dangerous'' situation: $\kappa(B) = 0.89$. A third one is within $D_{max}=30$ meters, but headed in a different direction, so it gets a mild criticality score $\kappa(B) = 0.60$. Instead, there are other less critical missed detections in the upper and lower parts of the image. These are farther than $D_{max}=30$ meters, and are headed in non-colliding trajectories: these are irrelevant, so they are worth $\kappa_d(B) = 0.00$.}

\begin{figure}[h!]
\centering
         \includegraphics[width=0.35\textwidth]{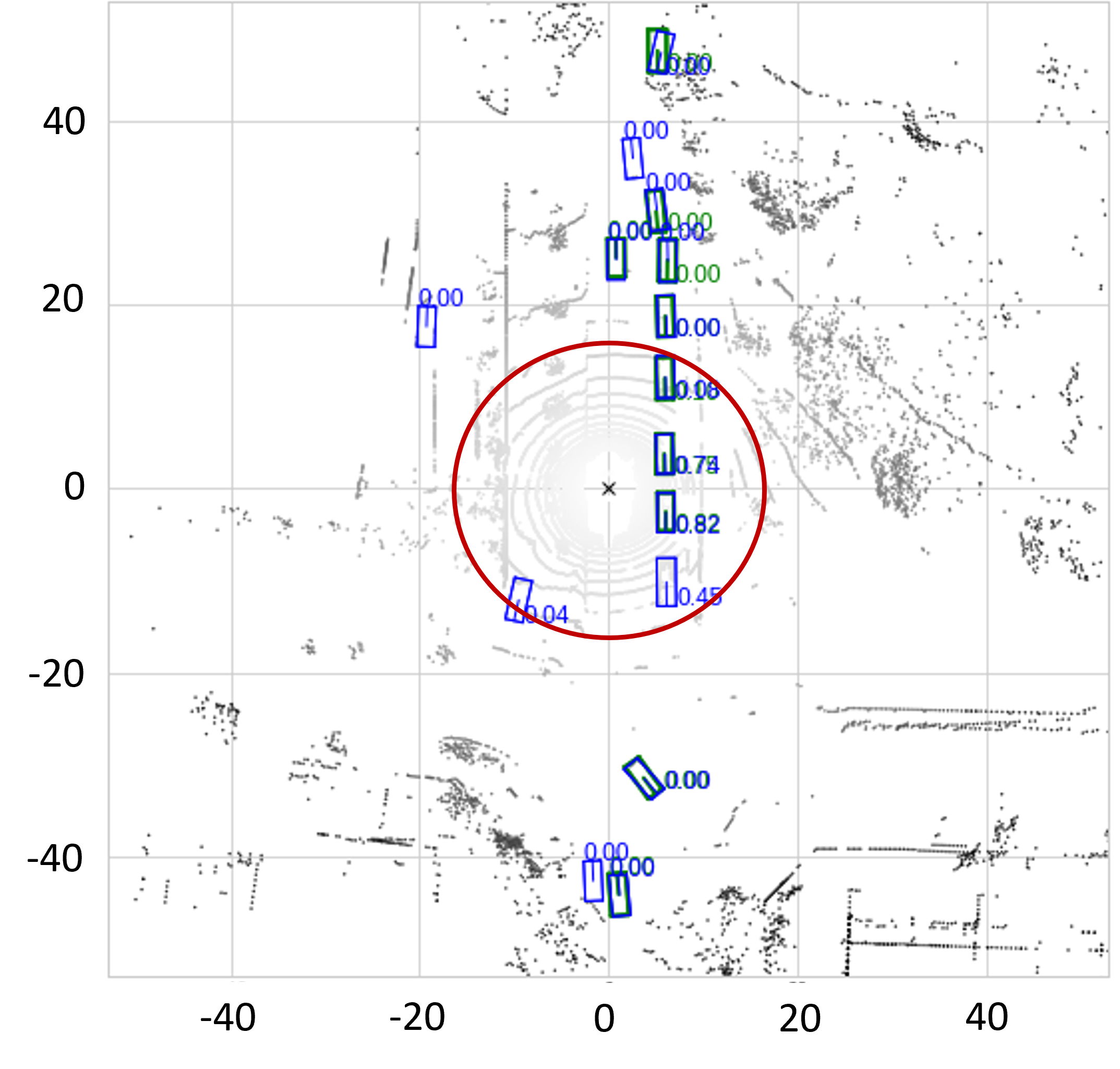}
         \caption{ $\kappa_d(B)$ computed for SEC with $D_{max}=15$, $R_{max}=20$, $T_{max}=10$. \textit{Best viewed in color.}}
         \label{fig:birdview6}
\end{figure}

\hl{Next, we explore the contribution of the distance criticality $\kappa_d(B)$. We consider SEC with $D_{max}=15$, $R_{max}=20$, $T_{max}=10$. Figure }\ref{fig:birdview6} \hl{shows that cars farther than $15$ meters from ego are assigned $\kappa_d(B)=0$; the closest to ego, the higher the $\kappa_d(B)$ values. The red circle is approximately $15$ meters radius: vehicles outside the circle have $\kappa_d(B)=0$.} 

\begin{figure}[h!]
\centering
         \includegraphics[width=0.35\textwidth]{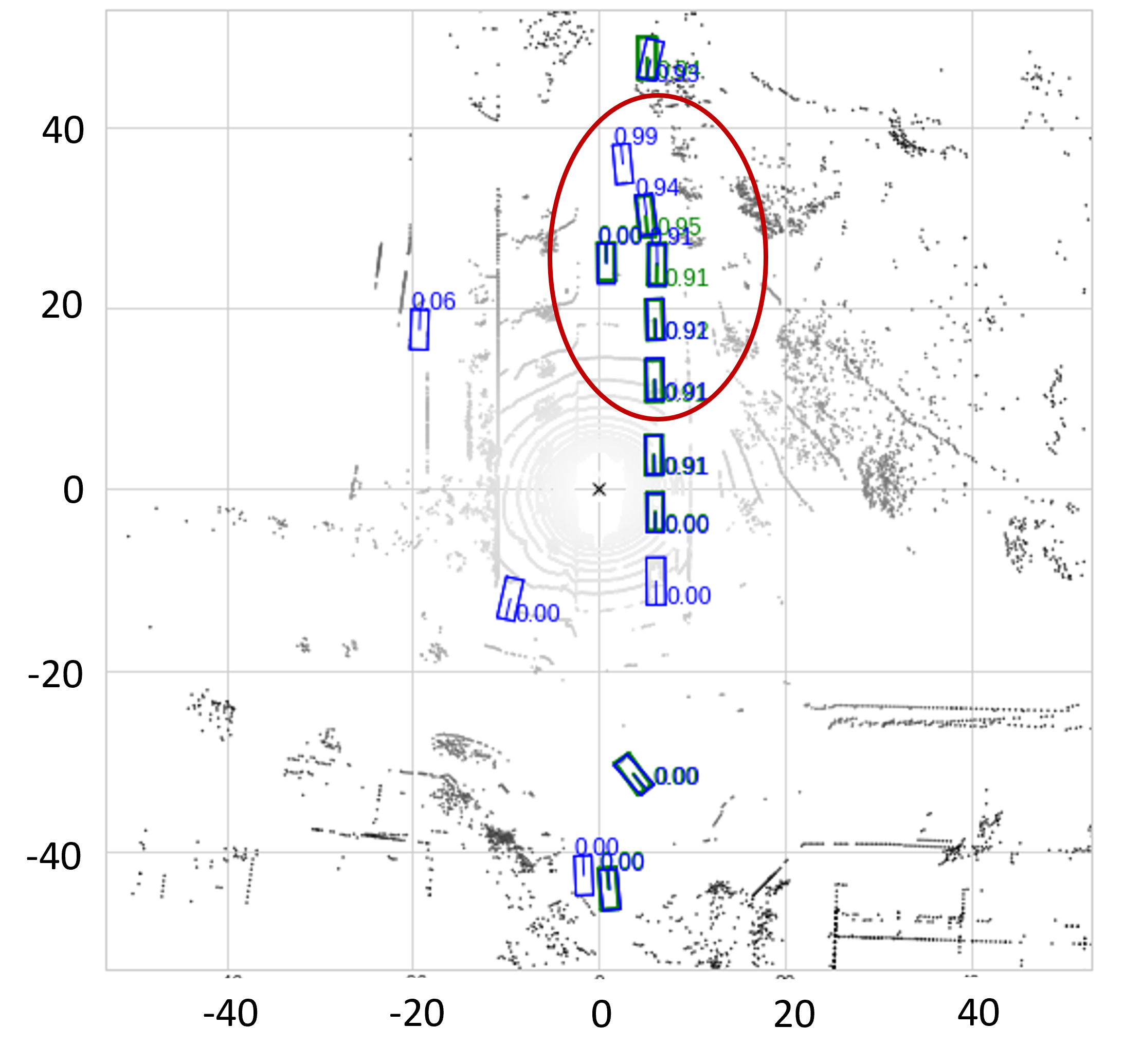}
         \caption{ $\kappa_r(B)$ computed for SEC with $D_{max}=15$, $R_{max}=20$, $T_{max}=10$. The red circle is an area of approximately $R_{max}=20$ meters from ego. \textit{Best viewed in color.}}
         \label{fig:birdview7}
\end{figure}

\hl{Figure }\ref{fig:birdview7} \hl{ shows the values of $\kappa_r(B)$ for the same scene and settings of Figure }\ref{fig:birdview6}\hl{. Cars with a trajectory passing closer to ego than $R_{max}=20$ are assigned $\kappa_r(B) > 0$; the closest to ego, the higher the values, meaning that those passing very close are at risk of collision.} \hl{Note that objects can be assigned a high $\kappa_r(B)$ value even if they appear to actually collide with ego. It is the case of the line of cars in the upper right part of Figure} \ref{fig:birdview7}\hl{, in the red circle: even if they are proceeding straight, they are passing very close on the right of ego, and thus the indicator of potential collision is close to 1.0.}

\hl{Similarly, Figure }\ref{fig:birdview8}\hl{ shows the values of $\kappa_t(B)$. Cars which may enter in a collision within $T=10$ seconds are assigned $\kappa_t(B) > 0$. The velocity of ego and each car is a determining factor to assign the criticality $\kappa_t(B)$: vehicles relatively close and in colliding trajectory may also have $\kappa_t(B)=0$ values if they are not expected to reach the collision point within $T=10$. In the red circle, there are opposite examples, of detected cars in colliding trajectories with ego but with $\kappa_t(B)=0$ and $\kappa_t(B)=0.96$ (note that this last one is a false positive). }

\section{Conclusions and Future Works}
\label{sec:conclusions}
 %\noindent
We argue that the most used measures for object detection 
% and their comparison
do not match the demands and peculiarities of a safety-critical system. Within the autonomous driving domain, currently adopted measures typically describe how good an object detector is at detecting \emph{all} the objects on the scene, while instead, for the purpose of an autonomous driving system, we are interested in detecting all the objects that will likely interfere with the driving task of the vehicle.

\hl{To this end, we show that 
the state-of-the-art evaluation of object detectors does not consider the possible role of the objects
in a specific scene, and in particular with respect to the driving
task of the vehicle performing the detection. }

\hl{Consequently,} we propose novel measures that take into account the concepts of
safety (detection of dangerous objects, which require immediate reaction, should be prioritized) and reliability (misdetections should not severely disrupt the continuity of the driving task). We build and exercise an object criticality model that performs a rating of the
objects, based on the distance from the subject vehicle, the possible colliding
trajectory, and the expected time to collision. 
Amongst the main results,
we show that our judgment on the performance of object detectors may be very
different when we consider the detection of i)  everything on the
scene (as it is usually done), or ii) only the relevant items. 
\hl{Depending on which of the two
cases is of interest, we may end up choosing different object detectors.
Further, we show that object detectors with high performance under case i)
can be less competitive in case ii), and vice-versa.} 

Last, an important implication of our object criticality model is that, when safety and reliability issues are considered, the selection of the most suitable object detector strictly depends on its desired use, i.e., on the requirements of the target application. Starting from application requirements, the desired configuration of our object criticality model is identified, measures are computed, and the most suitable object detector is selected.

\hl{Noteworthy, our analysis is not meant to prove that the evaluated object detectors are safe and reliable. Rather, it shows how the object criticality model allows establishing sound parameters that can be used to build, assess and tune object detectors for their application in safety-critical domains.}

\hl{We remark that object detection in complex
scenarios is still an open research topic that makes improvements every year }
\cite{8621614}\hl{, with new detectors that are proposed continuously; however,
defining new object detectors, or assessing the most up-to-date object
 detectors, is beyond the scope of this paper.}

%Future works will focus on considering alternatives to the
%metric
%model in Section \ref{sec:themodel}. Although the model in %this paper is fully applicable, it is also obvious that %more complex models could be
%elaborated, for example to represent the likelihood that %cars will steer or change speed. With the data available %on
%nuScenes, this is doable because trajectories of all %vehicles are traced, and they can be
%introduced in the model only at the cost of a more %complicated formulation to
%compute the intersection points. Further, we ignored %possible vertical offsets
%of objects, given the characteristics of the dataset %available; in our future
%works, we plan to also address these in the model so that %it can be used for
%datasets with vehicles on road slopes.
%\lm{Questo ultimo paragrafo lo toglierei. Invece di dire %che il modello fa caa', si potrebbe puntare sul fatto di %analizzare le varie configurazioni in scenari %reali/simulazione...}
%\lm{Si potrebbe anche dire che qualcuno (non di certo noi, almeno io non sono
%in grado :D) dovrà pensare a come ottimizzare i detector per queste metriche.}

\hl{As future work, we are currently working towards training an object detector whose goal is to maximise $AP_{crit}$. More precisely, the objective is to train to maximize a specific configuration of $R_S$ and $P_S$, rather than $R$ and $P$.} Intuitively, the object detector is intended to reward the detection of objects that are relevant (close and in colliding trajectories), and it is expected instead to be far less effective in the detection of objects that are not relevant for the driving tasks and that do not interfere with the elaboration of the trajectory of ego. \hl{Practically, this can be realized by a proper training phase, where the usual loss measurement approach is modified according to the principles and measures established in this work.}

\bibliographystyle{plain}
\bibliography{references}

\begin{IEEEbiography}[{\includegraphics[width=1in,height=1.25in,clip,keepaspectratio]{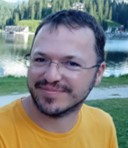}}]{Andrea Ceccarelli} is Associate Professor of Computer Science with the University of Florence, Florence, Italy. He holds a Ph.D. in
Informatics  and Automation Engineering (2012) from the University of Florence, Italy.  His primary research interests are in the design, monitoring and evaluation of dependable and secure systems, with a preference for experimental approaches. His scientific activities originated more than 100 papers that appeared in international conferences, workshops, and journals. He has been the Program Committee Co-Chair of the conferences SRDS and LADC. He is a member of the IFIP WG 10.4 on ''Dependable Computing and Fault-Tolerance''. 
\end{IEEEbiography}

\begin{IEEEbiography}[{\includegraphics[width=1in,height=1.25in,clip,keepaspectratio]{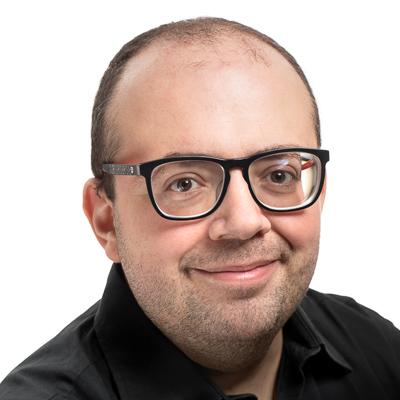}}]{Leonardo Montecchi} is Associate Professor at the Norwegian University of Science and Technology in Trondheim, Norway, since January 2022. Previously,
he was Assistant Professor at the University of
Campinas, Brazil (2017–2021). He holds a Ph.D. in
Computer Science, Systems and Telecommunications (2014) from the University of Florence, Italy,
where he also got his Bachelor's (2007) and Master's
(2010). His expertise revolves around modeling of
complex systems, including formal models, probabilistic models, and model-driven engineering. His
research focuses on modeling as a support to the Verification and Validation
of safety-critical and mission-critical systems. He regularly serves as reviewer
for international conferences and journals in the area of system and software
reliability. \end{IEEEbiography}
\clearpage

\EOD
\end{document}